\documentclass[review,5p,times,twocolumn,numbers]{elsarticle}
\usepackage[utf8]{inputenc}
\usepackage{comment}
\usepackage{tcolorbox}
\newtcolorbox{noteBox}{colback=orange!50,colframe=black!50}

\usepackage{amssymb}
\usepackage{amsmath}

\usepackage{url}

\usepackage[bookmarks=true]{hyperref}
\usepackage{algorithm2e}
\usepackage{graphicx}

\usepackage{float}
\usepackage{caption} 
\usepackage{enumitem}
\usepackage{adjustbox}
\usepackage{placeins}
\usepackage[dvipsnames]{xcolor}
\usepackage{tikz} 
\usetikzlibrary{shapes.geometric, arrows.meta, positioning}
\usepackage{booktabs} 
\usepackage{multirow}

\usepackage{cuted}
\usepackage{newfloat}
\DeclareFloatingEnvironment[fileext=lol]{codelisting}

\usepackage{listings}
\newcommand{\toggleBlock}[1]{}

\journal{The journal of systems and software}

\begin{document}

\begin{frontmatter}



\title{Modelling and Model-Checking a ROS2 Multi-Robot System using Timed Rebeca}


\author[mdu]{Hiep Hong Trinh\fnref{*}}
\fntext[*]{Corresponding author, email: hiep.hong.trinh@mdu.se}
\author[mdu,kth]{Marjan Sirjani}
\author[mdu]{Federico Ciccozzi}
\author[mdu]{Abu Naser Masud}
\author[mdu]{Mikael Sjödin} 

\affiliation[mdu]{organization={Mälardalen University},
            addressline={Universitetsplan 1, 721 23 Västerås}, 
            country={Sweden}}

\affiliation[kth]{organization={KTH Royal Institute of Technology},
            addressline={Brinellvägen 8, 114 28 Stockholm}, 
            country={Sweden}}

\begin{abstract}
Model-based development accelerates prototyping, enables earlier experimentation, and ensures rigorous validation of system de-
sign intents. In multi-agent systems with complex asynchronous interactions and concurrency, formal verification—particularly model-checking—offers an automated means of confirming that desired properties hold. Timed Rebeca, an actor-based modelling language supporting reactive, concurrent, and timed behaviors, together with its model-checking tool, provides a powerful framework for this purpose. By leveraging these capabilities, Timed Rebeca can intuitively capture ROS2 node graphs, recurring physical signals, motion primitives, and other time-convertible behaviors.

Nevertheless, modelling and verifying multi-robot systems entail significant challenges: abstracting intricate information, bridging the gap between discrete models and continuous system dynamics, and managing large state spaces while preserving fidelity. To address these challenges, we propose discretization strategies tailored to various data types and identify thresholds of abstraction that balance accuracy and tractability. We further introduce optimization techniques to accelerate verification.

Our work demonstrates how to systematically design and verify multi-robot systems through Timed Rebeca, efficiently transform continuous dynamics into discrete models for model-checking, and maintain a practical, bidirectional flow between the abstract model and the ROS2 implementation. The accompanying Rebeca and ROS2 codebases, made openly available, serve as a foundational reference for researchers and developers aiming to model and verify advanced autonomous robotic systems.
\end{abstract}



\begin{keyword}
multi-robot systems \sep modelling \sep model-checking \sep formal verification \sep ROS2 \sep Timed Rebeca
\end{keyword}

\end{frontmatter}



%
\section{Introduction}
System modelling offers a purposeful abstraction of a physical system that enables engineers to concentrate on its salient features, thereby facilitating the prediction of system qualities, the verification of specific properties, and communication with diverse stakeholders \cite{brown2004}. Within this framework, model-checking serves as an effective method for automatically verifying properties of finite-state concurrent systems \cite{ref8}, providing a deterministic outcome for each given scenario. Yet, the validity of model-based verification relies critically on the accuracy of the underlying model—a step commonly overlooked in practice.

Rebeca \cite{DBLP:journals/fuin/SirjaniMSB04,DBLP:conf/fmco/Sirjani06}
is an actor-based \cite{Agha1986} language specifically designed for modelling concurrent and reactive systems. Its extension, Timed Rebeca, adds temporal constructs that allow designers to handle timing semantics  \cite{DBLP:journals/scp/ReynissonSACJIS14,DBLP:conf/compsac/SirjaniLK20}. Offering a user-friendly syntax and an integrated model-checker with an IDE, Rebeca provides a comprehensive platform for building and verifying complex system designs \cite{DBLP:conf/birthday/Sirjani18}. Meanwhile, Robot Operating System (ROS), currently in its second generation (ROS2), is a well-established middleware for constructing robotic software \cite{ref11}. Ensuring functional correctness, safety, and reliability in multi-robot systems poses significant demands on both design and implementation. In such cases, modelling and model-checking can uncover concurrency issues early, substantially reducing overall development time and cost.

In this work, we leverage Timed Rebeca to create a holistic model of a multi-robot system and operational environment, complemented by equivalent ROS2 simulation code that preserves semantic alignment with the model. Using the Timed Rebeca model-checking tool, Afra \cite{DBLP:conf/fsen/KhamespanahSK23}, we systematically test behavioral algorithms, determine operational parameters and their respective safety thresholds, and verify properties such as goal reachability, deadlock freedom, and collision avoidance. Repeated simulations under analogous conditions in ROS2 confirmed the predictions of the model-checker, including rare safety issues that are not always evident through conventional simulation alone.

Using models for developing and verifying a multi-robot system introduces inherent challenges: selecting an appropriate level of abstraction for a complex setup, reconciling discrete model representations with continuous physical behaviors, and containing the computational explosion that can arise during verification. To meet these challenges, we propose different discretization strategies for different kinds of information and behaviours (e.g., map representation, robot movements, laser-based range detection), optimization techniques on discrete data, and confidence-based verification of semantic alignment between the model and its implementation counterpart. The novelty of this work is a pragmatic modelling approach that crafts a discrete model with good-enough abstraction level\footnote{Also called model resolution, granularity level, fine-grained level} for analyzing continuous robotic behaviors through model-checking rather than mere simulation, enabling verification over all possible executions. This is especially challenging in multi-robot systems, where concurrent autonomous behaviors cause state-space explosion. 

Compared to existing literature (see Section~\ref{sec:cmp}), our principal contributions and distinctions are:

\begin{enumerate} 
\item  Modelling the architecture of a representative multi-robot system employing Timed Rebeca, a timed actor-based language; and showcasing a round-trip engineering process that maintains semantic consistency between the abstract model and the deployed code (Sections \ref{sec:workflow}, ~\ref{sec:model}, and~\ref{sec:roscode}).

\item Developing strategies for mapping continuous physical behaviors to discrete formal models through diverse discretization techniques that effectively reduce the state space without compromising model fidelity. Combined with techniques for optimizing computations on discrete data, like the use of precomputed data and lookup tables, the model can be model-checked efficiently (Section~\ref{sec:challenge}).

\item Demonstrating a practical approach to model validation by comparing the structures and behaviours of the Timed Rebeca model and the ROS2 code, and ensuring that the model-based verification results align with the actual behaviors (Section \ref{sec:experiment}).
\end{enumerate}

The remainder of this paper is organized as follows: Section~\ref{sec:bg} lays out foundational concepts. Section~\ref{sec:design} describes the design and architecture of a multi-robot system using node graph in the context of ROS2. 
Section \ref{sec:workflow} explains our workflow, starting from node graphs and moving to ROS code and Rebeca model in an iterative and incremental way.
Section~\ref{sec:model} details our modelling approach with Timed Rebeca, while Section~\ref{sec:challenge} addresses challenges encountered during modelling and the solutions adopted. Section~\ref{sec:roscode} outlines the ROS2 code structure that mirrors the model. Section~\ref{sec:experiment} presents our experimental setup and results, and Section~\ref{sec:discuss} offers a broader discussion of the findings, limitations and technical soundness of our solution. After presenting the work, we compare it with others  in Section~\ref{sec:cmp}. Finally, Section~\ref{sec:conclude} concludes the paper.

\section{Background}\label{sec:bg}
\subsection{Robotics, Formal Verification, and Multi-Robot Systems}
This work originates from a master thesis project in \cite{ref1}
. 
Model-checking~\cite{ref15} is the most used technique because of its automation and similarity to exhaustive testing. Barriers of adoption include: 1) formal methods are perceived as a complicated extra step without adding final values, 2) lack of appropriate tools, 3) lack of interoperability between formalisms, 4) lack of clear guidance, and 5) difficulty in formalizing the step from a formal model to program code. The implementation code can be automatically generated or manually written to reflect the model; how to ensure that the code correctly represents the model is an open question. We consider these barriers in our work, in particular verifying the model's validity.
\toggleBlock{
Hugo A. et al. \cite{10.1145/3542945} performed a systematic literature review on techniques, processes, or tools for verifying system-level qualities of robotic and autonomous systems. Their summary shows widely used methods and tools are:  variants of temporal logics for modelling requirements and properties, variants of state-machines and transition systems for modelling system behaviors, Prism and Uppaal as model-checkers, Gazebo and Matlab/Simulink as simulators. Many papers lack rigorous measure of efficiency or adequacy for their proposed solutions and there is a little evidence of industrial applicability.
}

Obstacle detection and avoidance algorithms are the core components of multi-robot systems. These algorithms based on occupancy grid and A* path planning are studied in the context of Automated Guided Vehicle (AGV), which helps us to understand the navigational problem of real-world bulky mobile robots \cite{ref17}. Multi-robot systems should not be simply represented by dimensionless dots. The study on AGV motivates the need for verifying safety properties in multi-robot systems, especially in environments with human presence.
Formal models are used in multi-robot path planning problem, reinforcement learning techniques are used to synthesise robot behaviour, and model-checking and simulation techniques are used to evaluate task-level metrics of multi-robot performance. However, the success of these techniques is limited due to the model's accuracy, in particular the ability to capture and predict run-time behaviour. It has been suggested that future research should focus on developing smaller multi-robot models by imposing realistic assumptions over the sources of uncertainty and robot interactions so that they can be solved efficiently without sacrificing model accuracy~\cite{street2023multirobot}.

\subsection{ROS2}
\begin{figure}[tbp]
  \centering
  \includegraphics[width=0.5\textwidth]{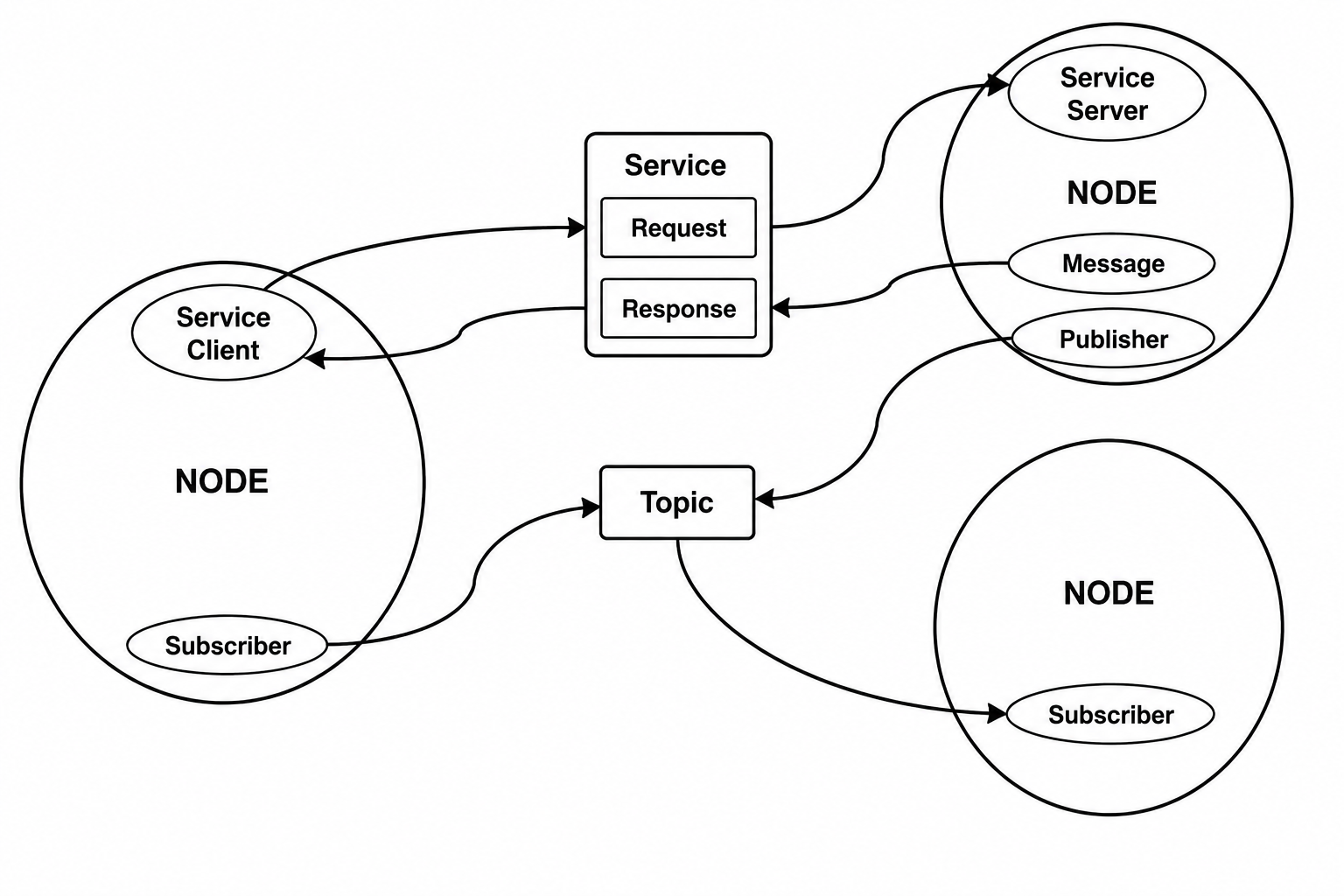}
  \caption{A ROS2 node graph (adapted from \url{docs.ros.org}).}
  \label{fig:ros2graph}
\end{figure}

A ROS2 robotic software system can be modelled as a network of distributed nodes which run concurrently and interact with each other through different inter-process asynchronous messaging mechanisms to achieve some robotic goals \cite{ref12}\cite{ref13}\cite{ref14}. ROS2 nodes do not share states, they only interact through messages. For each event, a callback function is defined to specify the behavior of the node when the event happens. Major communication mechanisms of ROS2 nodes are publish-subscribe topics, client-server services, and actions (see Fig.~\ref{fig:ros2graph}).

In this article, the term node is used in the broader distributed-system sense, referring to any computational entity in a distributed or multi-agent system. When such an entity is implemented in ROS2, it is specifically referred to as a ROS2 node.

\subsection{Actor Model and Timed Rebeca}
The actor model \cite{Agha1986} is a simple yet robust architecture for concurrent distributed systems. Each actor is a reactive unit with its own state, a queue of incoming messages to process, and a set of handlers that define how it should respond to each kind of message. Actors do not share states and only interact through asynchronous message passing. An actor can send messages to itself and other actors. Messages sent to an actor are put to its local queue and processed one by one in a run-to-completion manner. This way removes the complication in mutual exclusion of concurrent activities and basically eliminates deadlocks. The actor model is very suitable for modelling and implementing multiple autonomous agents systems.

Rebeca is an actor language for modelling and model-checking reactive, concurrent systems. The formal verification mechanisms are hidden underneath, modellers just need to focus on defining the structure and behaviour of the system through the high-level actor language. A model in Rebeca consists of environmental constants, blueprints of actor types (called \textit{reactiveclass}), and a \textit{main} function where instances of actors (called \textit{rebec}) are declared and activated. A reactive class contains declarations of state variables, known rebecs that it may interact with, message server methods ($msgsrv$) whose name, arguments and body map to type, parameters and the way a message is processed. The class can also contain ordinary methods for internal use only.


\begin{figure}[tbp]
\footnotesize
\begin{tabular}{lll}
Model & ::= & $(Class)^+\ Main$ \\
Main & ::= & \texttt{main} $\{InstanceDecl^*\}$ \\
InstanceDecl & ::= & $C\ r\ (\langle r\rangle^*) : (\langle c\rangle^*)$ \\
Class & ::= & \texttt{reactiveclass} $C\ \{KnownRebecs\ Vars\ MsgSrv^*\}$ \\
KnowRebecs & ::= & \texttt{knownrebecs} $\{VarDcl\}$ \\
Vars & ::= & \texttt{statevars} $\{VarDcl\}$ \\
VarDcl & ::= & $\langle T\ v\rangle;$ \\
MsgSrv & ::= & \texttt{msgsrv} $m(VarDcl)\ \{Stmt^*\}$ \\
Stmt & ::= & $v=e;$ $\mid$ Call $\mid$ \texttt{if}$(e)$ MSt [\texttt{else} MSt] $\mid$ \texttt{delay}$(t);$ \\
Call & ::= & $r.m(\langle e\rangle^*)$ [\texttt{deadline} $e$] [\texttt{after} $e$] \\
MSt & ::= & $\{Stmt^*\}$ $\mid$ Stmt \\
\end{tabular}
\caption{Abstract syntax of Timed Rebeca. Angle brackets ⟨⟩ denotes meta
parenthesis, superscripts + and {*} respectively are used for repetition of one or
more and repetition of zero or more times. Combination of with repetition
is used for comma separated list. Brackets [] are used for optional syntax.
Identifiers C, T, m, v, c, e, and r respectively denote class, type, method name,
variable, constant, expressions, and rebec name, respectively.}
\label{fig:trsyntax}
\end{figure}

Extract \ref{fig:trsyntax} shows the syntax of Timed Rebeca with three language primitives to augment message processing with time attributes \cite{DBLP:journals/scp/ReynissonSACJIS14,DBLP:journals/corr/abs-2309-07302}:
\begin{itemize}
    \item \textbf{delay(t)}: to model the advancement of time during message execution in a \textit{msgsvr},
    \item \textbf{deadline(t)}: to specify expiry time of a message (the timeout to handle the event), e.g.  \texttt{r1.doLaserScan()} \texttt{deadline(100)},
    \item \textbf{after(t)}: to model time of a future event or message delivery time. Periodic events are modelled by a self-looping message server with a fixed repeat period, while discrete events of variable scheduled times are modelled with a dynamically computed period, e.g., \texttt{r1.updateMovingStatus()\ after(time\_taken)}.    
\end{itemize}
Each actor has a logical clock to order messages by their logical timestamps and all logical clocks are assumed to be fully synchronized. This assumption is independent of physical clocks, which can be unsynchronized and need to map to a shared timeline as in the model. A Timed Rebeca model can be equivalently converted to a set of timed automata \cite{ehsanphd}. Properties can be defined as assertions over state variables. 
The model defined in Timed Rebeca is compiled and checked by the built-in model checker to verify desired properties. The support of actor model, time semantics and language friendliness \cite{DBLP:conf/birthday/Sirjani18} reasons our selection of Timed Rebeca\footnote{Herefrom we use Rebeca and Timed Rebeca interchangeably.} for modelling autonomous robot systems.

\subsection{Discretization}
Model-checking relies on a discrete finite-state model for exhaustive exploration of the state space (that's why state variables have to be bounded enumerable types like $integer$, $boolean$ or fixed-size arrays of such), while dynamics like movements in a robot system are often continuous, expressed by $real-number$ variables. When mapping a continuous system to a discrete model, we perform a process called discretization~\cite{fishwick2007}, which involves sampling the continuous behavior at selected intervals and approximating it with representative values. The resulting discrete model preserves the relevant system dynamics within an acceptable level of accuracy.
Discrete abstraction allows modelling a continuous system as a discrete-time or discrete-state system, which can be more easily simulated and analyzed using discrete mathematics and computational techniques.

\section{Designing a Multi-Robot System using a Node Graph}\label{sec:design}
To demonstrate our approach, we need a multi-robot system to work with. However, since available systems often deal with only one robot or contain many unnecessary features, we have to create an exemplary one to gain full control. Finally, we designed a hypothetical multi-robot system, modelled it in Timed Rebeca, and created corresponding ROS2 simulation code\footnote{Full modelling code: \url{https://github.com/thhiep/ros2rebeca_model}}(\textit{the code} in the remainder of the paper).

A multiple autonomous mobile robots (AMR) system was chosen for the following reasons: 1) we can project the robots and obstacles onto a 2D map to track their movements and interactions, and handling 2D geometry is less complicated than 3D (like with robotic arms), provided limited mathematical facilities of modelling languages; 2) a mobile robot is equal to a mobile obstacle, thus the multi-robot problem subsumes the single-robot one which requires modelling other non-static obstacles (like a wandering human); 3) obstacle detection, path finding, collision avoidance are typical robotic problems to deal with and successful modelling practices with these can be generalized and inherited for several other cases; and 4) with multiple robots moving in the same area, the need of guaranteeing collision and deadlock freedom is more demanding, therefore model-checking is more valuable.

For effective modelling and simulation, it is essential to develop a thorough understanding of the problem domain, specifically in the context of robotics and multi-robot systems. Without it, there is a risk of either oversimplifying or overcomplicating the scenarios being addressed. This section provides a detailed explanation of the structure and operation of a system involving multiple mobile robots.

\subsection{Basic Structure}
A multi-AMR system can be designed with variations in architectural topography and communication flows, but consists of the following core elements:
\begin{itemize}[topsep=0pt]
    \item A floor map: which is a 2D projection of static obstacles on the floor plane. The map is treated as a known environmental knowledge to the robots.

    \item Robot nodes: each robot has an ID, a footprint shape, a current pose $(x,y,\theta)$, linear velocity $\overrightarrow{v}$ and angular velocity $\overrightarrow{\omega}$, a queue of targets (intermediate and final) to reach, a planned path trajectory to walk, and a set of responses that defines how it should react to different events. Each robot reports its status to a map server node for system-wide observation and supervision.

    \item Map server node: a central coordinating node that keeps track of the current positions of the robots on the map, provides common services, and checks system-wide properties during execution.
\end{itemize}
The physical dimensions of a robot must be considered as they affect algorithms for collision detection and path planning. As a rectangular shape is more representative for vehicles, robot shapes are generalized to rectangular profiles in the model.

\subsection{Obstacle Detection}\label{sec:eq2}
\begin{figure}[tbp]
    \centering
    \includegraphics[width=0.5\textwidth]{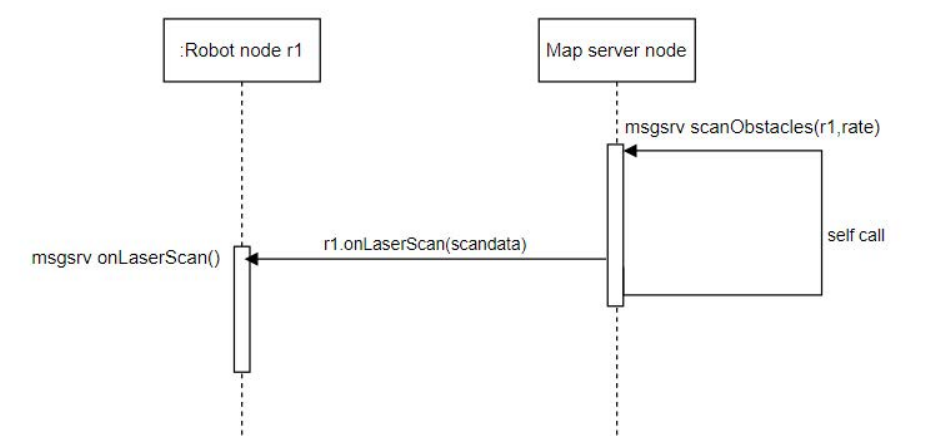}
    \caption{Modelling laser scan events}
    \label{fig:laserscan}
\end{figure}

LIDAR sensors are extensively used for range detection \cite{ref21}. In ROS2, laser scan data is provided by the physical sensor and encapsulated in a standard message format\footnote{\url{https://docs.ros2.org/foxy/api/sensor\_msgs/msg/LaserScan.html}}. In modelling and simulation, this data is generated programmatically. In the discrete model, the generation is done with discrete intervals (refer to \ref{laserscandiscrete}) and the periodic recurrence is simulated through a timed loop (see Fig. \ref{fig:laserscan}).
The robot maintains a matrix of the nearest obstacles, which it uses to identify obstructions in a given direction. This matrix is updated with each newly scanned signal; however, as the robot moves, the stored obstacle data becomes outdated since the robot is no longer in the same position as when the last scan was received. Consequently, there is a critical safety relationship between the robot's movement speed and the scanning rate, ensuring the robot does not move faster than the rate at which obstacle data is updated. This relationship is analogous to the Nyquist-Shannon theorem \cite{ref18}, which states: \textit{The sampling frequency of a signal must be greater than twice the highest frequency of the input signal}. In this context, the time a robot takes to perform an action should be at least twice the period of the signal it relies upon, i.e.,
\begin{equation} 
    \label{eq:2}    
    \mathit{action\_time} > 2 \ast \mathit{signal\_period}\qquad
\end{equation}

\subsection{Collision Detection}\label{coldef}
\begin{figure}[htbp]
    \centering
    \includegraphics[width=0.5\textwidth]{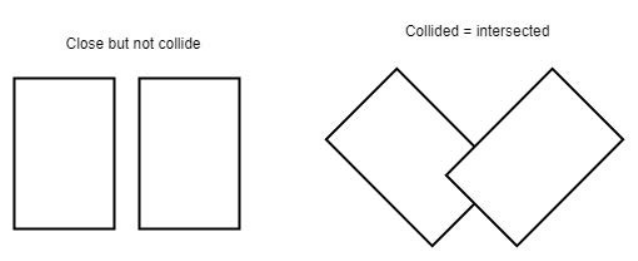}
    \caption{Collision check based on intersection of shadows}
    \label{fig:collide}
\end{figure}

In the real world, a physical collision occurs without defining how it happens logically. However, in modelling and simulation, it is necessary to define the conditions under which a software-detected collision occurs. For round robots, collision detection can be simplified by measuring the distance between the centers of two robots. For rectangular robots, this method is insufficient, as they may be positioned parallel and close to one another without any actual collision. In general, a collision is detected by checking for an intersection between the bounding areas (shadows) of two robots (see Fig. \ref{fig:collide}). Based on each robot's pose and dimensions, we can calculate its current boundary, extended by a safety margin, and determine whether it intersects with the extended boundary of another robot.

\subsection{Navigational Problems}
A mobile robot typically utilizes a floor map to navigate. In ROS2, the map data is saved to a  portable gray map (PGM) image file, where the gray level of each pixel depicts the occupancy status of a point. The map file is then parsed into an occupancy grid (with resolution defined in a YAML file) to facilitate graph-based pathfinding algorithms, of which A*\cite{ref19, ref19b} and its variants are widely used. Pathfinding is an optimization problem with no guaranteed shortcut to a predictable solution. Therefore, in modelling, we must implement a pathfinding algorithm to generate walkable paths and ensure semantic alignment between the model and the code (that's why we opted not to use an existing library like Nav2). However, the path is collision-free only at the time of generation and is not future-proof, as mobile obstacles may appear or move unexpectedly.
\begin{figure}[H]
    \centering    
    \includegraphics[scale=0.4]{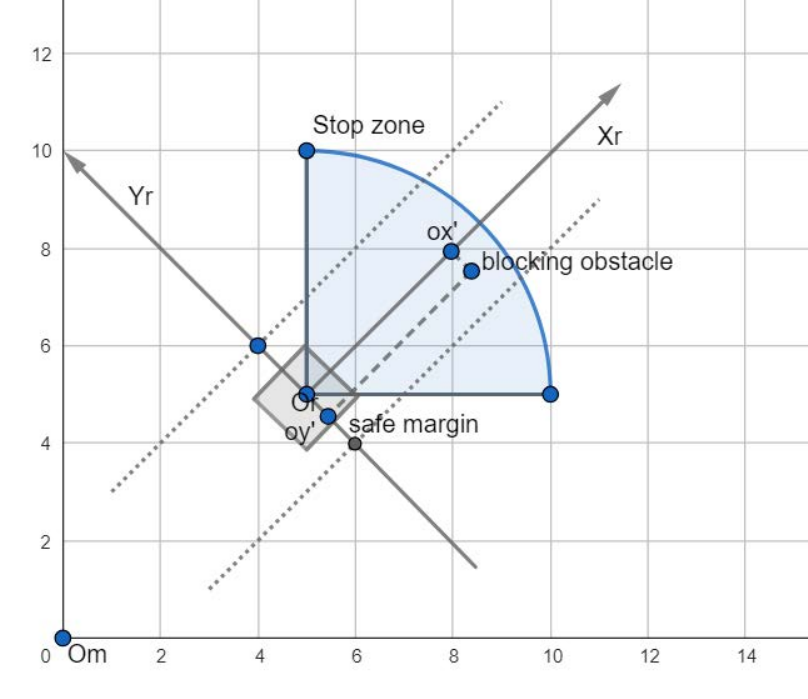}    
    \caption{Blocking obstacles}
    \label{fig:blocking}
\end{figure}
When the robot encounters obstacles along its route, it must decide whether to stop and wait, continue moving, or reroute to avoid obstacles or clear congestion. We developed a behavioral algorithm in Alg.\ref{alg:algo1} inspired by how humans walk or drive in real life. The algorithm follows a \textit{look-think-act} sequence. Obstacles are categorized as either blocking or non-blocking. A blocking obstacle lies directly in the robot's path, while a non-blocking obstacle is outside of the robot’s immediate frontal area (see Fig. \ref{fig:blocking}). Even if an obstacle is close to the robot, it is considered non-blocking if it does not impede forward movement. The available frontal space is determined by estimating the distance to the nearest object in the robot's current direction. The robot then uses a heuristic value to assess whether there is sufficient space to proceed; otherwise, it stops and waits. This heuristic is influenced by factors such as inertia, speed, and braking distance. However, for simplicity in the modelling process, this value is set as a constant ($\textsc{stop\_zone}$).

\RestyleAlgo{ruled}
\begin{algorithm}[h]
\small
\caption{Human-like movement algorithm}\label{alg:algo1}
\SetAlgoLined
\SetKwProg{proc}{procedure}{ begin}{end}
\proc{moveNext(P, micro\_step, wait\_inc)}{
\nl Let $w_{0}$ be the current position of the robot

\nl Let $P = \{w_1,\ldots,w_n\}$ be the generated path where $w_1$ is the next waypoint and $w_n$ is the goal

\nl \If{(P is empty)}{   {// final waypoint reached}\;
    Stop(); return;
}

\nl First, rotate the robot to the next waypoint $w_{1}$ so that it looks towards that way (skip if already) 

\nl Estimate $d$ = distance to the nearest blocking obstacle in that direction

\nl Let $h$ = some heuristic value of required frontal space to be able to move forward safely

\nl \eIf{(d $>$ h)}{
//move one microstep forward, reset waiting status\;
    $wait\_count \gets 0$\;
    $w_{0} \gets w_{0} + micro\_step$\;
    \If{(Robot has reached $w_{1}$, meaning $w_{0} \equiv w_{1}$)} {Remove $w_{1}$ from stack $P$}
    $moveNext(P,micro\_step,wait\_inc)$
}{
    {Stop(); Wait($wait\_inc$);\;$wait\_count \gets wait\_count + wait\_inc$}\;    
    \eIf{(wait\_count \textgreater max\_waiting\_time)}{ 
       {Resolve congestion}       
    }{
      {// continue waiting or moving}
      $moveNext(P,micro\_step, wait\_inc)$
    }
}
}
\end{algorithm}%

\RestyleAlgo{ruled}
\begin{algorithm}
\small
\SetAlgoLined
\caption{Congestion resolving algorithm}\label{alg:algo2}
\nl Let $getObstacleAtDir(dir)$ be the function for estimating distance to the nearest blocking obstacle in a direction $dir$

\nl Let $backdirs$ be the set of directions on the opposite side of the current direction

\nl Let $od$ be a list of $\{dir,freespace\}$ records

\nl Estimate free space in all back directions:\\
$foreach(dir \; in\; backdirs)$\\$\;\;od.push(\{dir,getObstacleAtDir(dir)\});$

\nl Sort $od$ by descending order of $freespace$ field

\nl Pick a random direction among the top (e.g. 3) group
\\Let $selected = random(1,max(count(od),3))-1$
\\Let $backdir = od[selected].dir;$

\nl Pick a random number of possible backoff steps in the selected direction (e.g. 3 to 10 steps):
\\Let$\;backsteps = random(3, max(od[selected].freespace,10));$

\nl Walk the robot back to the "$backdir$" direction by "$backsteps$" steps, then request a new path from the new location to the destination

\end{algorithm}

\section{Modelling and Development  Workflow}\label{sec:workflow}
\begin{figure}[tbp]
    \centering
    \includegraphics[width=0.5\textwidth]{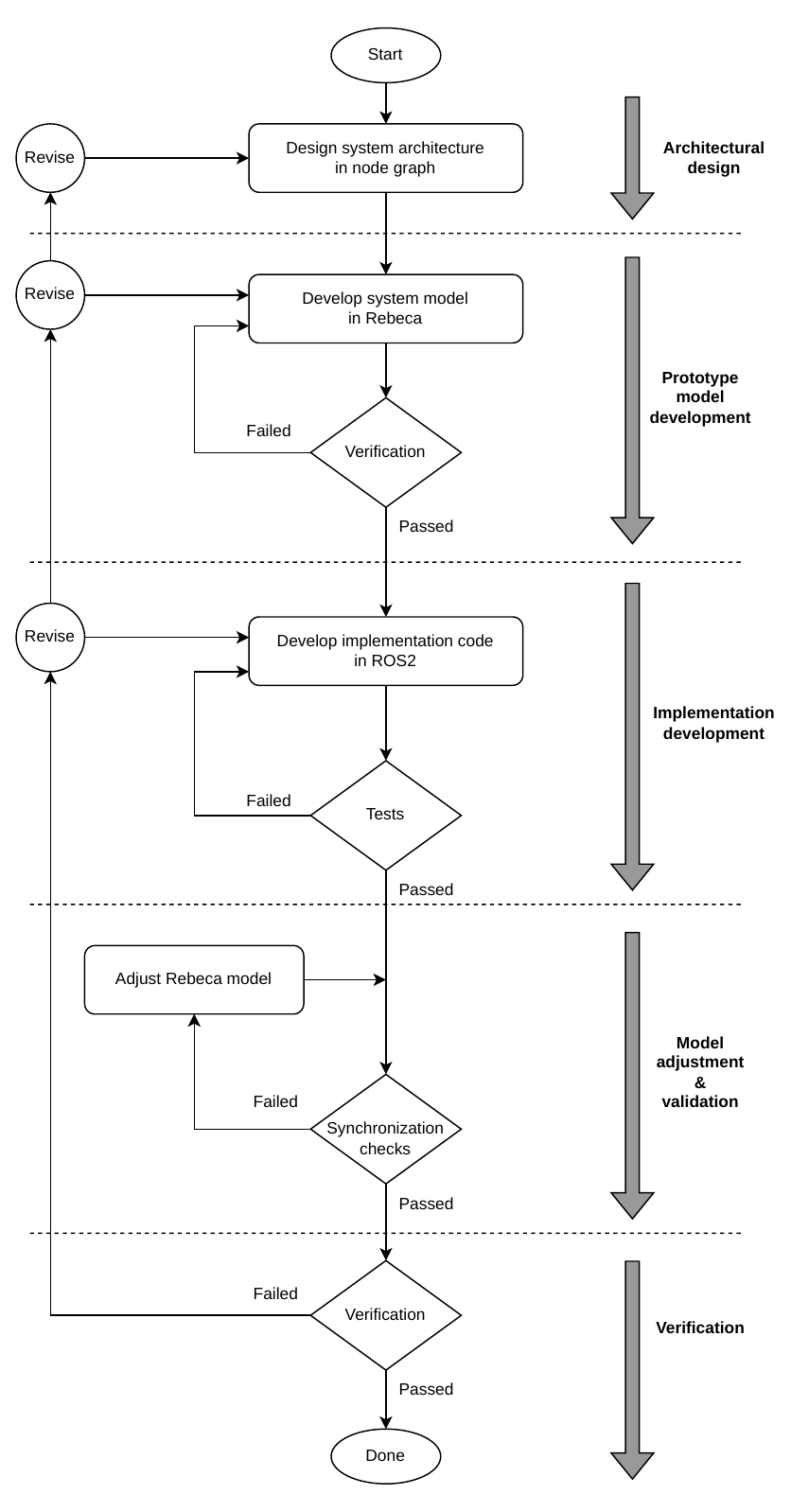}
    \caption{
   The workflow for designing, modelling, implementation and verification}
    \label{fig:workflow}
\end{figure}

\begingroup
The modelling and development workflow of our multi-robot system is depicted in Fig.~\ref{fig:workflow}.
In the first phase, we select a robotic problem to model, and design the architecture as a node graph. Next, we develop Timed Rebeca model based on the node architecture. We use a few scenarios which we expected to pass the model checking, for example, a simple scenario including only a single robot or scenarios with multiple robots with easy navigation paths from source to destination with low possibility of collisions. In the third phase, we develop the ROS2 code based on the node graph and the Rebeca model. Since the Rebeca model lacks visualization, some issues are identified during ROS2 execution with visualizers. Examples of these issues include using not optimized paths by the robots, or avoiding obstacles and rerouting too early. 
 Modifications are made to the ROS2 code to remove such problems. In the fourth phase, the changes are propagated back to the model to ensure semantic consistency between the model and the ROS2 code.
 We also calibrated the model in terms of granularity level until it behaves the same as the code under selected scenarios. If we observe  the Timed Rebeca model behaving differently compared to the ROS2 code, we need to revise the model again. For example, when we check the paths or when we observe a collision or deadlock in the simulation but model checking did not detect it, we conclude that the precision of the model is not enough and we increase the granularity.

The model is validated when it passes adequate synchronization checks, in which we compare the model and the implementation code in terms of structural 
and behavioral equivalence. Structural equivalence is based on the node graph of both the model and the code, and behavioral equivalence includes same information flow, same behavioral algorithms (path finding, movement, obstacle avoidance and congestion resolution), same generated paths and final outcome like lack of deadlock and collisions, under same settings of robot and map configuration. The ROS2 visualization and the traces in the state space generated by Afra IDE are used for comparing paths.

In the final phase, we perform the verification of the ROS2 code by model checking the Timed Rebeca model and verifying safety properties under different scenarios. If problems are detected, the loop can be restarted to modify the model, the code, or the design of the system.
\endgroup

\section{Modelling the System in Timed Rebeca}\label{sec:model}

\subsection{General Mapping Rules}
ROS2 nodes are equivalent to reactive objects (\emph{rebecs}) in Rebeca. Service requests and responses in ROS2 are mapped to Rebeca messages and message servers with discretized and bounded arguments. Timers and periodic signals are modelled using timed loops in Rebeca via the $\mathit{after}()$ primitive. Real-number variables are subject to different discretization mechanisms in the model. General mapping rules are summarized in Table \ref{table:mapping}, items marked with (*) are discretized and explained in the next sections. We can see that all robotic information types are captured in Rebeca with equivalent or abstracted representation.

\begin{table}[H]
\centering
\caption{Robotic information in ROS2 application and in Rebeca model}
\label{table:mapping}
\resizebox{1.0\columnwidth}{!}{%
\begin{tabular}{|p{3cm}|p{4cm}|p{4cm}|}
\hline
\textbf{Information type} & \textbf{In ROS2} & \textbf{In Rebeca} \\ \hline
Parameters & Constants in launch file & $env$ constants \\ \hline
Floor map & PGM and YAML files & Equivalent occupation grid (*) \\ \hline
Node definition & ROS2 node class & Reactive class \\ \hline
Node instance & Instance of ROS2 node class & Instance of reactive class ($rebec$) \\ \hline
Event type & ROS2 message structure & $msgsvr$ name and formal arguments (*) \\ \hline
Event instance & ROS2 message & $msgsvr$ actual arguments (*) \\ \hline
Event response & ROS2 callback function & $msgsvr$ body \\ \hline 
Pub-sub channel & ROS2 pub-sub topic & Data publisher is a $msgsvr$ of the service provider $rebec$. Data subscriber is a callback $msgsvr$ in the service consumer $rebec$. \\ \hline 

Robot physical shape & URDF file & Shape of shadow (rectangular) \\ \hline
Robot pose & ROS2 odometry message & Grid cell and direction $(i,j,rdir)$ (*) \\ \hline
Range detection & ROS2 laser scan messages & Simulation of scanning result (*)  \\ \hline
Movement primitives  & Periodic micro-step updates (5-10ms) & Next pose scheduling with varying $\Delta t$ (*) \\ \hline
Speed & Travelled distance in a micro-step & Time taken to transit to next state, $\Delta t$  \\ \hline
Path finding & A* algorithm & Same \\ \hline
Obstacle avoidance & Look-think-act algorithm & Same \\ \hline
Congestion resolution & Random backoff & Same \\ \hline
\end{tabular}}
\end{table}
\hfill

\subsection{Modelling State Evolution Over Time}
For each robot, the progression of states over time is modelled as a sequence of timed events as outlined in Alg.\ref{alg:algo3}. Based on the current state and external events (e.g., the arrival of a new laser scan or a new path), the next state and its occurence time are computed. State transitions and time advancement are simulated by using a time specifier in Rebeca, typically the $\mathit{after}()$ primitive. 

\begin{algorithm}[tbp]
\small
\SetAlgoLined
\caption{Time-wise discretization of state}\label{alg:algo3}
\nl Let $loc_1$ be the current pose (state) of the robot

\nl Let $loc_2$ be the next pose of the robot (if the robot is waiting, $loc_2$ is same as $loc_1$)

\nl Let $\Delta t$ be the time taken to switch to $loc_2$ 

\nl Then: \texttt{setNextState}($loc_2$) \texttt{after}($\Delta t$)
\end{algorithm}

The transition time, denoted as $\Delta t$, is calculated based on the duration a robot takes to either:
\begin{itemize}
    \item change its direction: $\Delta t = |\theta_2 - \theta_1| / \omega$, where $\omega$ is rotation speed, $\theta_1, \theta_2$ are current and next angles,
    \item move to an adjacent cell: $\Delta t = s / v$ where $v$ is linear speed, $s=\text{\texttt{CELL\_WIDTH}}$ if moving along the edge, or $s=\text{\texttt{CELL\_WIDTH}}*\sqrt{2}$ if diagonally, or 
    \item continue waiting: $\Delta t =$ \texttt{waiting\_increment}.
\end{itemize}

In simulation, the robot's movement is updated by small steps (every 5-10 miliseconds) to simulate the smooth continuous movement in reality, while in the model, it is done by discrete macro-steps since it only "hops" from cell to cell or rotates by a multiple of $45^0$. This approach allows the conversion of velocities into time, enabling the use of Rebeca's time semantics to model the timing sequence of the movement updates accurately. The time-wise discretization of states enables capturing only meaningful moments, skipping unnecessary iterations in the middle -- which are typically found in a tick-based time model, causing state explosion and inefficiency.

\subsection{Overall Interaction Flows and the Model's Code}
\begin{figure}[tbp]
    \centering
    \includegraphics[scale=0.75]{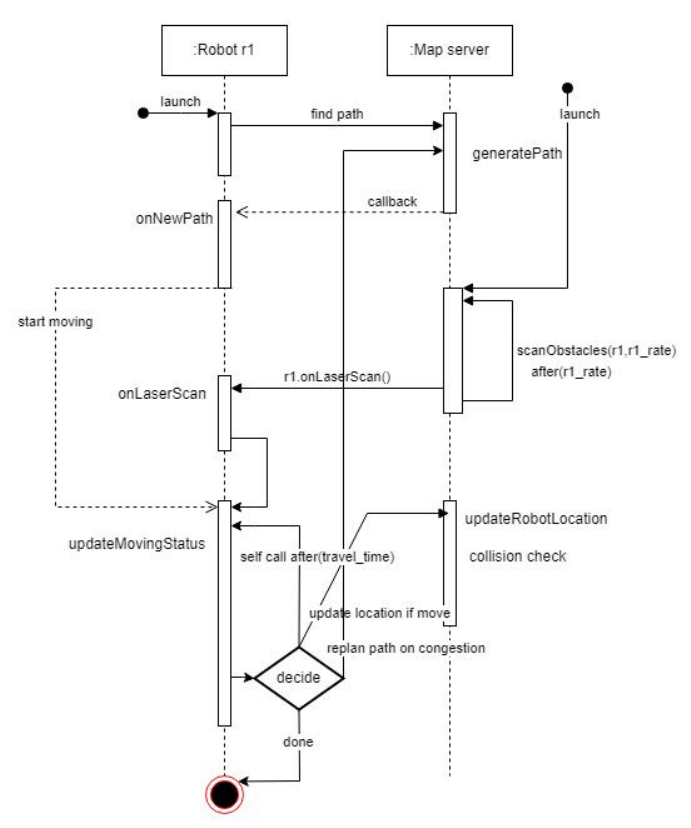}
    \caption{Sequence diagram of map server–robot interactions}
    \label{fig:map-robot-seq}
\end{figure}

Fig.~\ref{fig:map-robot-seq} shows the interactions between a robot node and the map server node. The map server node exposes the following services: \texttt{scanObstacles} (generating LIDAR sensory data), $\mathit{generatePath}$ (applying A* based algorithm for finding paths), \texttt{updateRobotLocation} (updating robots' locations on the map). Each robot node implements the following callbacks: \texttt{onNewPath} (start moving on the generated path), \texttt{onLaserScan} (updating the matrix representing the nearest obstacles and decide to continue moving or stop) and \texttt{updateMovingStatus} (deciding in a loop whether a robot can make a move, stop, wait, or reroute).

The full modelling code is provided on GitHub, here we only present its basic structure in Listing ~\ref{fig:mcode}. In the Timed Rebeca model, first the configuration parameters are defined as $env$ constants, and then two reactive classes are defined to model the mobile robots (class $Node$) and the coordination center (class $MapServer$).  The interactions of these two classes is visualized in Fig.\ref{fig:map-robot-seq}. In the $main\{\}$ part in the code, the map server and the robots  are instantiated. The parameters of the robots denote  the ID, starting point, target point, tolerance distance to target, linear speed, waiting time, and a $scanFirst$ flag telling whether it should always wait for a new obstacle detection scan before moving. 
At start, each robot will send \texttt{theMap} a \texttt{generatePath()} request to create a path to its target. After receiving the path, it will start planning the moves along the computed waypoints. Depending on the situation it has to cope with the obstacles on its path, each robot will decide what to do next - continue moving, stop and wait, or reroute. For rerouting the robot asks the map server for a new path.
The model execution stops when no more state evolution is possible, which happens when the destination is reached or a deadlock/livelock is detected.




{
\captionof{codelisting}{Modelling code in Timed Rebeca}
\label{fig:mcode}
\begin{lstlisting}[language=Java,basicstyle=\ttfamily\footnotesize]
env int MAP_WIDTH = 50;
env int MAP_HEIGHT = 50;
env double MAP_RESOL = 0.25500;
env double CELL_WIDTH = MAP_RESOL;
env double CELL_DIAG = SQRT2 * MAP_RESOL;
env int XMAX = MAP_HEIGHT - 1;
env int YMAX = MAP_WIDTH - 1;

env double ROBOT_WIDTH = 0.4;
env double ROBOT_LENGTH = 0.4;
env double SAFE_MARGIN = 0.02;
env double STOP_ZONE = 0.3;
env double COLLISION_RADIUS = 0.30;
env double COLLISION_MARGIN = 0.02;

env double ROBOT_ANGULAR_VELOCITY = 360;
env int ANGLE_INC = 2;
env int FIELD_OF_VIEW = 180;
env int LASERSCAN_RATE = 100;
env double LASERSCAN_MAX_DISTANCE = 10.0;

env boolean COMBINE_ZICZAC = true;
env int BUFFER_SIZE = 101;

reactiveclass Node(50) {

    knownrebecs { MapServer theMap; }

    statevars {
        byte rindex;
        int targetX, targetY;
        int rx, ry, rdir, velocity;
        int distance2target;
        int target_tolerance;

        int[101] obstacles;
        int front_obstacle;

        int[11] targets;
        int[101] moves;

        int movedIdx;
        boolean isWaiting;
        int waits;
        int max_waiting_rounds;
        int failures;
        boolean scan_before_move;

        int default_velocity;
        int default_time_cross;
        int default_time_cross_diag;

        int safe_front;
        int safe_side;
    }

    Node(byte idx, int x, int y, int tx, int ty, int tolerance, double real_velocity, int maxwait, boolean scanfirst) {
        // initialization routine
    }

    @priority(3)
    msgsrv onLaserScan(int[101] scandata, int[101] angles) {

        if (isWaiting ||
           (!scan_before_move && distance2target > target_tolerance))
            updateMovingStatus(!scan_before_move);
    }

    @priority(1)
    msgsrv updateMovingStatus(boolean moveNext) {
        // move / stop / wait decision
    }

    @priority(10)
    msgsrv onNewPath(int[101] path) {
        // handle new path
    }
}

reactiveclass MapServer(50) {

    knownrebecs {
        Node r1; Node r2; Node r3; Node r4; Node r5;
    }

    statevars {
        int[21] rloc;
        int collisions;        
    }

    MapServer(int rcount) {
        collisions = 0;
        
        //initialize robot information array rloc
        
        //start laser scan simulation events...
    }

    @priority(1)
    msgsrv scanObstacles(Node robot, int rindex, int rate, int fov, int max_distance) {

        int[101] scandata;

        robot.onLaserScan(scandata, angles) deadline(rate + 5);

        scanObstacles(robot, rindex, rate, fov, max_distance) after(rate);
    }

    @priority(5)
    msgsrv updateRobotLocation(int idx, int rx, int ry, int rdir, int velocity) {
        // update location + collision check
    }

    @priority(3)
    msgsrv generatePath(Node robot, int robot_index, int startX, int startY, int targetX, int targetY) {
        int[101] path;

        // A* path generation
        // zigzag optimization

        robot.onNewPath(path);
    }
}

main {
    @priority(3)
    Node r1(theMap) : (1,5,5,45,45,2,0.5,1500,false);

    @priority(5)
    Node r2(theMap) : (2,5,15,45,35,2,0.7,2000,false);

    @priority(6)
    Node r3(theMap) : (3,5,25,45,25,2,0.8,2500,false);

    @priority(4)
    Node r4(theMap) : (4,5,35,45,15,2,0.5,3000,false);

    @priority(2)
    Node r5(theMap) : (5,5,45,45,5,2,0.3,1500,false);

    @priority(1)
    MapServer theMap(r1,r2,r3,r4,r5) : (5);
}
\end{lstlisting}
}
\leavevmode\indent

\subsection{Correctness Properties for Model Checking}
For a multi-AMR system,  we aim to ensure the following properties: 
\begin{enumerate}
\item Collision freedom: At all times, no two robots collide. 
\item Reachability (and deadlock freedom): Eventually, all robots reach their destinations.
\item Congestion freedom: The robots do not get stuck or indefinitely blocked.
\end{enumerate}
These properties are formally defined as follows: 

\textbf{Collision freedom:} Collisions between robots are detected when the map server node receives and processes robot location updates (\textit{updateRobotLocation}). The abstracted poses of the robots in the discrete world are converted back to the continuous world to detect collision using the mechanism defined in Sec. \ref{coldef}, if a collision is detected, the \textit{collisions} counter is incremented.
Collisions with static obstacles are already avoided via the path generation algorithm by keeping the robot at a safe distance from the obstacles. Formally, collision freedom is expressed by the following CTL formula:
\[
\mathit{collisionFree} := \mathrm {AG}(\mathit{theMap}.\mathit{collisions} = 0)
\]

\textbf{Reachability:} Each robot is assigned a tolerance for reaching its target. A state variable ($distance2target$) tracks the remaining octile distance to the target cell. A robot $r$ is considered to have reached its destination, $done_r=true$, when it is within the specified tolerance of the target (as implemented in ROS2) and has no further planned moves that drive it beyond the target. So, this property is defined as:
\[
\mathit{finallyDone} := \mathrm {EF}(\forall r: \mathit{done_r}=\mathit{true})
\]

\textbf{Deadlock freedom:} Deadlocks are automatically detected by the model checker \cite{10.1145/2414639.2414645} when the system reaches a state in which the robots are unable to proceed, preventing further progress towards a final state. Livelock is identified by the \mbox{\textit{failures}} counter that tracks the number of unsuccessful rerouting attempts made by a robot. If the value of this counter exceeds a predefined threshold, the robot is considered to be potentially stuck. This property is expressed as
\[
\mathit{liveLockFree} := \mathrm {AG}(\forall r:r.\mathit{failures} \le \mathit{tolerance_{failures}}) 
\]

In Timed Rebeca, temporal logic operators are equivalently converted to assertions as follows:
\begin{itemize}
    \item $\mathrm {AG}(a)$, meaning $a$ must hold true any time in any execution path, is equivalent to a native assertion expression $p:=a$.
    \item $\mathrm {EF}(a)$, meaning finally $a$ becomes true in at least one execution path. To verify that all robots finally arrive at their destinations, we add a counter assertion \mbox{$\mathit{notDone}:=\neg(\forall r: \mathit{done_r=true})$}, which will stop the modelchecker whenever all robots arrive. Violating this assertion means model-checking has passed successfully. Note that the robots are bounded by the collision free and livelock free conditions, which will prevent them from moving around endlessly without reaching targets.
\end{itemize}
Listing ~\ref{fig:properties} shows the property specification file, where predicates are defined in the \textit{define} block and properties are specified in the \textit{assertion} block.

{
\captionof{codelisting}{Property definition in Rebeca}
\label{fig:properties}
\begin{lstlisting}[language=Java,basicstyle=\ttfamily\footnotesize,breaklines=true,breakatwhitespace=false,columns=fullflexible]
property{
	define {
		done1 = (r1.distance2target <=r1.target_tolerance) 
			&& (r1.moves[0]==0 || (r1.moves[0]>0 && r1.moves[0]-r1.moveidx<=r1.target_tolerance));
		
		done2 = (r2.distance2target <=r2.target_tolerance) 
			&& (r2.moves[0]==0 || (r2.moves[0]>0 && r2.moves[0]-r2.moveidx<=r2.target_tolerance));
		
		done3 = (r3.distance2target <=r3.target_tolerance) 
			&& (r3.moves[0]==0 || (r3.moves[0]>0 && r3.moves[0]-r3.moveidx<=r3.target_tolerance));
		
		done4 = (r4.distance2target <=r4.target_tolerance) 
			&& (r4.moves[0]==0 || (r4.moves[0]>0 && r4.moves[0]-r4.moveidx<=r4.target_tolerance));
		
		done5 = (r5.distance2target <=r5.target_tolerance) 
			&& (r5.moves[0]==0 || (r5.moves[0]>0 && r5.moves[0]-r5.moveidx<=r5.target_tolerance));

		noCollisions = theMap.collisions==0;
		
		noLivelock = r1.failures<=2 && r2.failures<=2 && r3.failures<=2 && r4.failures<=2 && r5.failures<=2;		
	}
	
	Assertion{
		notDone: !(done1 && done2 && done3 && done4 && done5);
		collisionFree: noCollisions;
		liveLockFree: noLivelock;
	}
}
\end{lstlisting}
}
\leavevmode\indent

\section{Modelling Challenges and Strategies}\label{sec:challenge}
Rigorous verification of robotic systems is known as a complex, interdisciplinary task while formal methods are often hindered by the state-space explosion problem \cite{DBLP:journals/firai/FoughaliZ22}.   
Modelling robotic behaviors for the purpose of model-checking presents significant challenges. What are the factors -- 'the Bad' and 'the Ugly' -- that hinder or complicate the path toward successful modelling?

\subsection{The Bad: Discretization Strategies and Pitfalls}\label{sec:eq4}
\FloatBarrier
The primary issue, 'the Bad,' lies in the inherent gap between a discrete model and a continuous system. This gap persists, regardless of the specific approximation or sampling technique used. A key challenge in this context is the trade-off between increasing the granularity of the model and managing the resulting state space explosion, a common and critical problem in model-checking \cite{ref8}.

\begin{figure}[H]
    \centering
    \includegraphics[scale=.9]{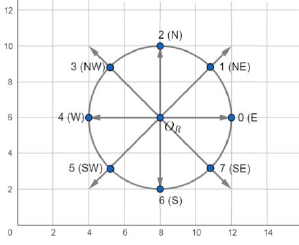}
    \caption{Discretization of robot directions}
    \label{fig:directions}
\end{figure}

\textbf{Discretizing robot pose:}
As the obstacle map is discretized to an occupancy grid, in the model a robot can only be at one cell and move to an adjacent cell in 8 possible directions. These 8 angles are multipliers of $45^o$, which is convenient for trigonometric computations (see Fig.~\ref{fig:directions}). Using this strategy, we mapped the continuous pose $(x,y,\theta)$ from the real system to a discrete pose $(rx,ry,rdir)$ in the model, preserving essential information while ensuring the pose remains sufficiently enumerable for model-checking purposes.
So, from map coordinate to grid:
\[
\begin{array}{lll}
rx &=& \mathit{floor}(\frac{x} { \textsc{cell\_width}});\\			
ry &=& \mathit{floor}(\frac{y} { \textsc{cell\_width}});\\
rdir &=& \mathit{floor}(\frac{\theta}{45});
\end{array}
\]
From grid to map coordinate, the robot is conventionally placed at the centre of the cell:
\[
\begin{array}{lll}
x &=& (rx + 0.5) * \textsc{cell\_width};\\		
y &=& (ry + 0.5) * \textsc{cell\_width};\\
\theta&=&rdir*45;
\end{array}
\]

\textbf{Discretizing laser beam angles:}\label{laserscandiscrete} In reality, a LIDAR sensor's angular resolution (beam step) can be a small fraction (e.g. $0.02^o$). In the model, it is discretized to $2^o$ ($1$ is too narrow, resulting in overlapped scans, while $3$ is too wide, resulting in skipping adjacent objects); field of view is set to a multiplier of beam step, typically $180^o$, $270^o$ or $360^o$.

\begin{figure}[H]
    \centering    
    \includegraphics[scale=0.4]{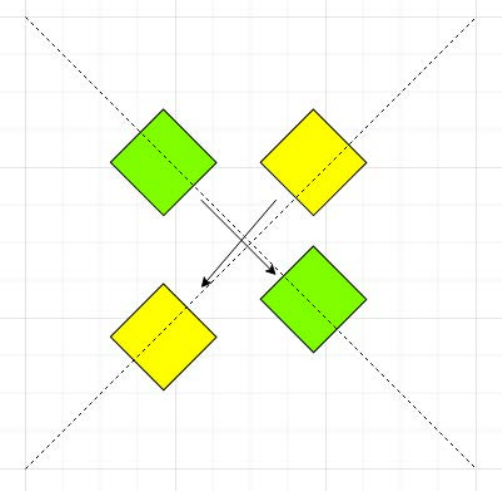}
    \caption{Grid resolution vs. robot size}
    \label{fig:stepsize}
\end{figure}

\textbf{Bridging the gap:}
When discretizing a continuous measurement, there is a risk of information loss between two consecutive steps, which increases as the step size becomes larger. This issue was observed in relation to the grid resolution and the robot size (see Fig.~\ref{fig:stepsize}). In the model, each movement step corresponds to a full cell edge ($\textsc{cell\_width}$) or diagonal ($\textsc{cell\_width}*\sqrt{2}$). If the step length exceeds the robot’s physical size, potential collisions occurring between steps may be overlooked. To mitigate this, the step length must be smaller than the robot's size, ensuring that any mid-step collisions are accurately detected due to the overlap between consecutive steps:
\begin{equation}
\label{eq:4} 
\texttt{cell\_width}< \texttt{robot\_length}/\sqrt{2}\quad
\end{equation}
If the step size is too small, the state space grows exponentially without further gain in accuracy. Determining the 'enough' threshold is therefore crucial in balancing between correctness and computational efficiency of the model.

\subsection{The Ugly: Workarounds and Optimization Techniques} 
Modelling robotic systems require both high-level and low-level expressiveness power of the modelling language and platform to capture system architecture and to do unavoidable computations and algorithms same as in implementation (e.g. coordinate translation, path finding). However, formal modelling languages are intended for specifying high-level logic rather than handling low-level computations. There are unmatched programming facilities in Rebeca that need to be coped with\footnote{For the version in use at the time this work was done}:
\begin{itemize}
    \item Inadequate type system: Rebeca does not support complex collection and structured types, which are neccessary for describing robotic data. Arrays can only be of static size.
    \item Lack of common mathematical, IO and utility functions as well as mechanism for using externally defined ones. Modellers therefore have to recreate their own implementation of very basic functions and cannot reuse existing computation libraries.
    \item Limited debugging, visualizing capabilities.
\end{itemize}
To overcome these language-specific inconveniences and to boost computational efficiency in model-checking, we employ several techniques, described in the following.

\textbf{Helper script:} We developed a script in PHP to assist with the testing and troubleshooting of the modelling code: 1) prepare static data: convert between PNG and PGM image file formats, generate occupancy grid matrix and precompute trigonometric values; 2) write script code in Rebeca style to test and debug; 3) visualize map state to troubleshoot algorithms.

\textbf{Precomputed data and lookup tables:} The occupancy grid corresponding to a PGM map file is parsed by the script and saved to a text file. When this data is needed in the model, it is copied to a local variable. As angles (robot angles, laser beam angles) in the model were discretized to integers, we use the script also to precompute $cos,sin,tan$ of all angles from $0^o$ to $360^o$, saving to a key-value array to look up later. As a result, on-the-fly trigonometric calculations are eliminated, significantly reducing computation time. This optimization is particularly impactful in model-checking, where the same calculations are repeatedly executed during each transition.

\section{Developing ROS2 Code from the Model}\label{sec:roscode}

After having a working prototype model, we manually\footnote{As an initial step, we chose not to automate the model-to-code translation due to the need to uncover various underlying issues. Automation is planned for future iterations.} develop the corresponding ROS2 code\footnote{Full ROS2 code: \url{https://github.com/thhiep/ros2rebeca_code}
}, adhering to the established mapping rules and maintaining the same node structure, interaction flows, and behavioral algorithms as defined in the model.
The code runs on ROS2 Foxy version, Ubuntu 20.04.5 LTS on a VirtualBox virtual machine. RViz2 was chosen as the visualizer, instead of Gazebo, because its 2D visualization is lightweight. Gazebo is often used in parallel to broadcast $/laserscan$ topic for RViz2 to capture; here we simulate it inside the code to have a self-contained, compact solution. 
\begin{table}[tbp]
\centering
\caption{File structure of the ROS2 demo codebase}
\label{table:ros2code}
\resizebox{1.0\columnwidth}{!}{%
\begin{tabular}{l}
    \includegraphics[width=0.5\textwidth]{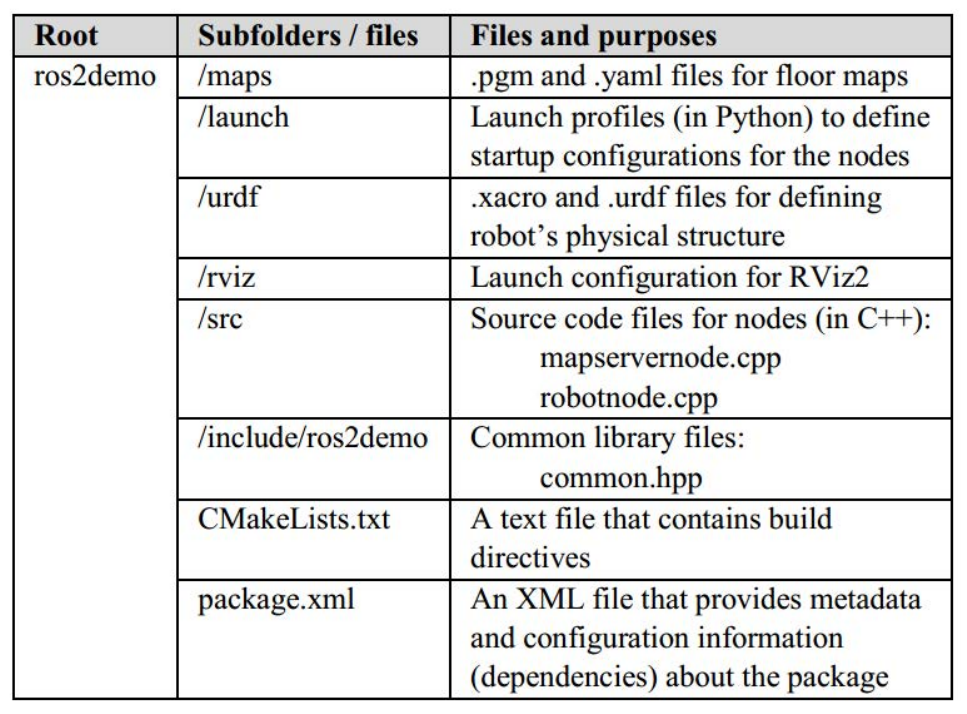}
\end{tabular}%
}
\end{table}
Table \ref{table:ros2code} presents the structure of the ROS2 codebase. The code implements two kinds of nodes, as in the model: map server node and robot nodes. The ROS2 nodes need to generate and broadcast data in standard topics so that RViz2 can capture and visualize on the graphical interface (including occupancy grid, robot state, standard transformations, laser scan, path, and additional markers).

\textbf{Multi-robot handling:} Most open-source ROS2 codebases are designed for single-robot systems. To enable multi-robot functionality, each robot is assigned a unique identifier (e.g., $r1$, $r2$) which also serves as its namespace. This ID is appended—either automatically by ROS2 or manually—to each robot’s interfaces and frames (e.g., $/r2/laserscan$). A XACRO template is used to define each robot’s parameters, such as name, dimensions, and color, to generate a unique URDF model for each robot. Shared interfaces and frames (e.g., $/map$), common to all robots, remain unprefixed.

\textbf{Change synchronization:} In the program, generated paths are optimized to be shorter and more efficient. This is done through \textit{waypoint consolidation} (combining waypoints that lie in the same direction to reduce loop iterations and increase speed) and \textit{zig-zag point elimination} (skipping zig-zag points along  diagonals, allowing for straighter and faster diagonal movement). The model is then updated to align with these optimizations introduced in the program.

\section{Simulation and Model Checking Results}\label{sec:experiment}

We developed the Timed Rebeca model and the corresponding ROS2 code following the modeling workflow described in Sec.~\ref{sec:workflow}.
To enable a meaningful comparison between model verification and execution,  same configuration parameters were used for both the ROS2 experiments and the model-checking setup. 

A randomly generated 50x50 grid map is used, with the resolution set to 0.255 to fit RViz2 viewport. The robot dimensions are set to 0.4x0.4, which satisfies the fine-grained threshold in Equation \ref{eq:4} (see Section~\ref{sec:eq4}). An extreme scenario is designed with five robots that are required to navigate intersecting paths to reach their respective targets across the map. This scenario creates a high risk of collisions and congestion, allowing us to rigorously evaluate the correctness properties and assess the effectiveness of the behavioral algorithms. In this scenario, robots may start at arbitrary initial positions and are assigned arbitrary goals. Waypoints are computed by the A*-based algorithm. Due to the concurrent execution of the robots, the system behavior is non-deterministic.

The experimental parameters and statistical counters are explained in Table~\ref{table:legend}.
\bgroup
\def\arraystretch{1.5}
\begin{table}[tbp]
\centering
\caption{Explanation of parameters and counters}
\label{table:legend}
\resizebox{1.0\columnwidth}{!}{%
\begin{tabular}{|p{3cm}|p{8cm}|}
\hline
\textbf{Name} & \textbf{Meaning}                                        \\ \hline
Scan rate           & Laser scan rate in miliseconds                          \\ \hline
Speed limit & Threshold for linear speed, bounded by safety relationship with scan rate                          \\ \hline
Stop zone   & A heuristic value the robot uses to decide to stop up to estimated distance to blocking obstacles \\ \hline
Speed               & Linear movement speed (m/s)                                            \\ \hline
Wait        & Maximum waiting time in miliseconds before the robot decides to back off and find another path               \\ \hline
Analysis result     & Model-checking result                                   \\ \hline
States              & Number of states traversed by the modelchecker          \\ \hline
Transitions         & Number of state transitions tried by the modelchecker   \\ \hline
Simulations         & Number of ROS2 code executions under the same settings \\ \hline
Simulation results  & Results of ROS2 code executions                       \\ \hline
\end{tabular}}
\end{table}
\egroup

We present two types of scenarios, \textit{\textbf{working and non-working,}} each illustrated with three representative examples. For each type of scenario, we provide a table of the corresponding experimental parameters and statistics, along with screenshots from the simulation. These examples demonstrate that the issues predicted by model checking (such as collisions or congestion) indeed occur in the ROS2 executions.

\begin{table}[H]
\centering
\caption{\centering{The results for ROS execution and Timed Rebeca model checking for the cases which works properly and satisfied the correctness properties.}}
\label{table:3}
\resizebox{1.0\columnwidth}{!}{%
\begin{tabular}{l}
    \includegraphics[scale=1.2]{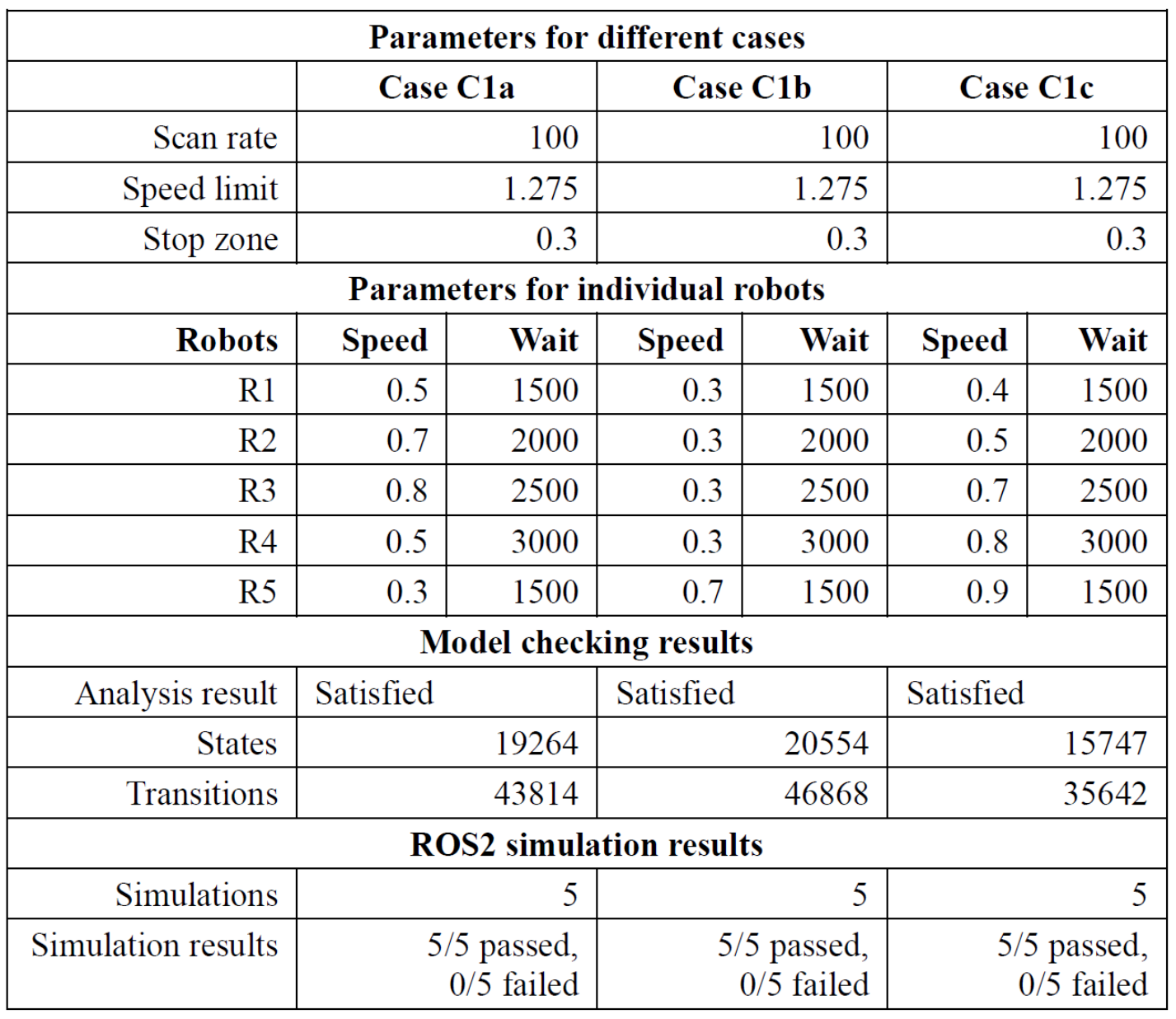}
\end{tabular}%
}
\end{table}

\textbf{Working Cases (Cases C1a, C1b, C1c).} Table \ref{table:3} provides the statistical and experimental parameters and results for cases C1a, C1b, C1c. All three cases share the same global safety envelope (scan rate, speed limit, stop zone). The speed limit is set to 1.275 m/s, satisfying the timing constraints from Equation~\ref{eq:2}, which is formally derived from the sensor scan rate ($100ms$) via an analogy to the Nyquist-Shannon sampling theorem. Just as a signal must be sampled at more than twice its highest frequency to be faithfully reconstructed, any robot action must take longer than twice the scan period to be reliably observable. Since the scan period is $100ms$, edge traversal must take more than $200ms$, which constrains the maximum safe speed to $0.255m / 0.2s = 1.275 m/s$. Exceeding this limit means the robot can cross an entire grid cell between two consecutive scans, making its position unobservable at cell-level granularity and invalidating the discrete model upon which the formal verification relies. The working cases (C1a, C1b, C1c) all respect this bound, which is why the model checking results carry over faithfully to the ROS2 simulations.

What varies is how individual robot speeds and wait times are assigned across the five robots. This variation in individual robot behavior contributes to the success of these cases by naturally staggering movements and reducing the likelihood of simultaneous intersection conflicts.
Model checking returns \mbox{\textbf{Satisfied}} in all three cases after exhaustively exploring their respective state spaces. The differences in state/transition counts reflect varying degrees of non-determinism — more varied speed combinations produce more interleavings for the checker to explore. All 5/5 ROS2 simulations then confirm what the model proves: all robots reach their destinations without problems. 

\begin{figure}[tbp]
    \centering
    \includegraphics[width=\columnwidth,keepaspectratio]{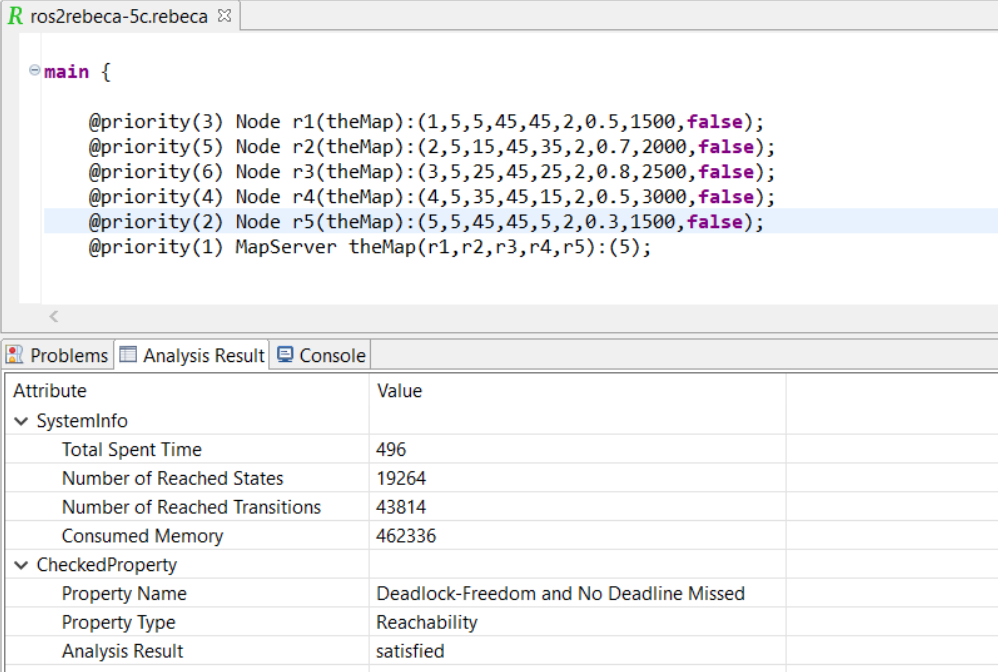}
\includegraphics[width=\columnwidth,keepaspectratio]{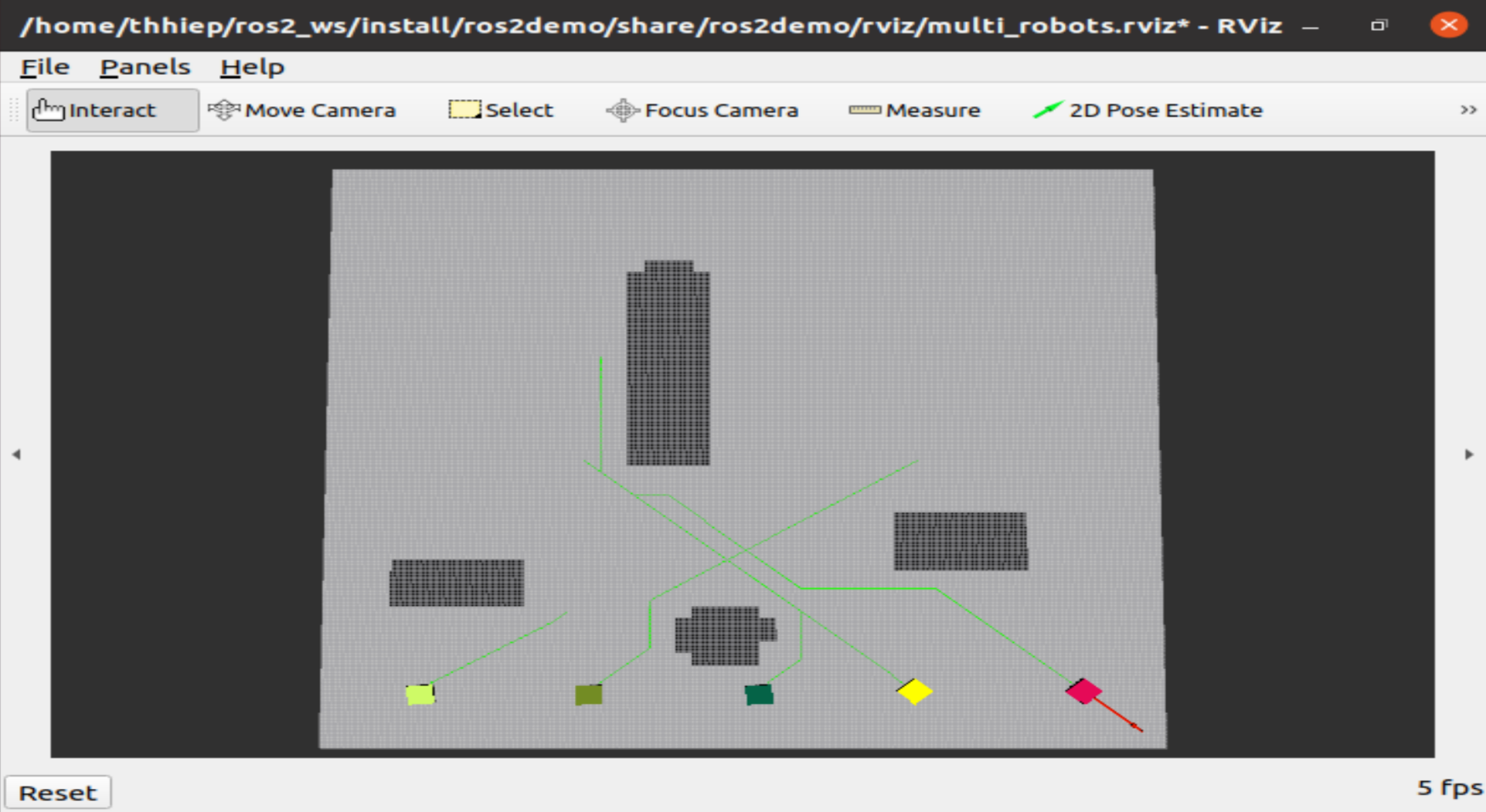}
    \caption{Screenshots of Afra results (up) and ROS2 simulation results (down) for a working case}
    \label{fig:working}
\end{figure}
Figure~\ref{fig:working} illustrates Case C1a as a representative of working cases. The upper panel shows the Timed Rebeca model instantiation, where five robots are assigned distinct priorities, speeds, and wait times. The middle panel shows the model checker output: after exploring $19,264$ states and $43,814$ transitions in $496ms$, deadlock freedom, no missed deadline, and reachability properties are verified as satisfied. The lower panel shows the corresponding ROS2 visualization in RViz2, where the green lines trace the A*-computed paths of the five robots navigating around obstacles toward their respective goals. The crossing paths represent the deliberately congested scenario, yet all robots successfully reach their targets without collision or deadlock, consistent with the model checking result.

\begin{table}[H]
\centering
\caption{ Non-working cases}
\label{table:4}
\resizebox{1.0\columnwidth}{!}{%
\begin{tabular}{l}
    \includegraphics[scale=1.2]{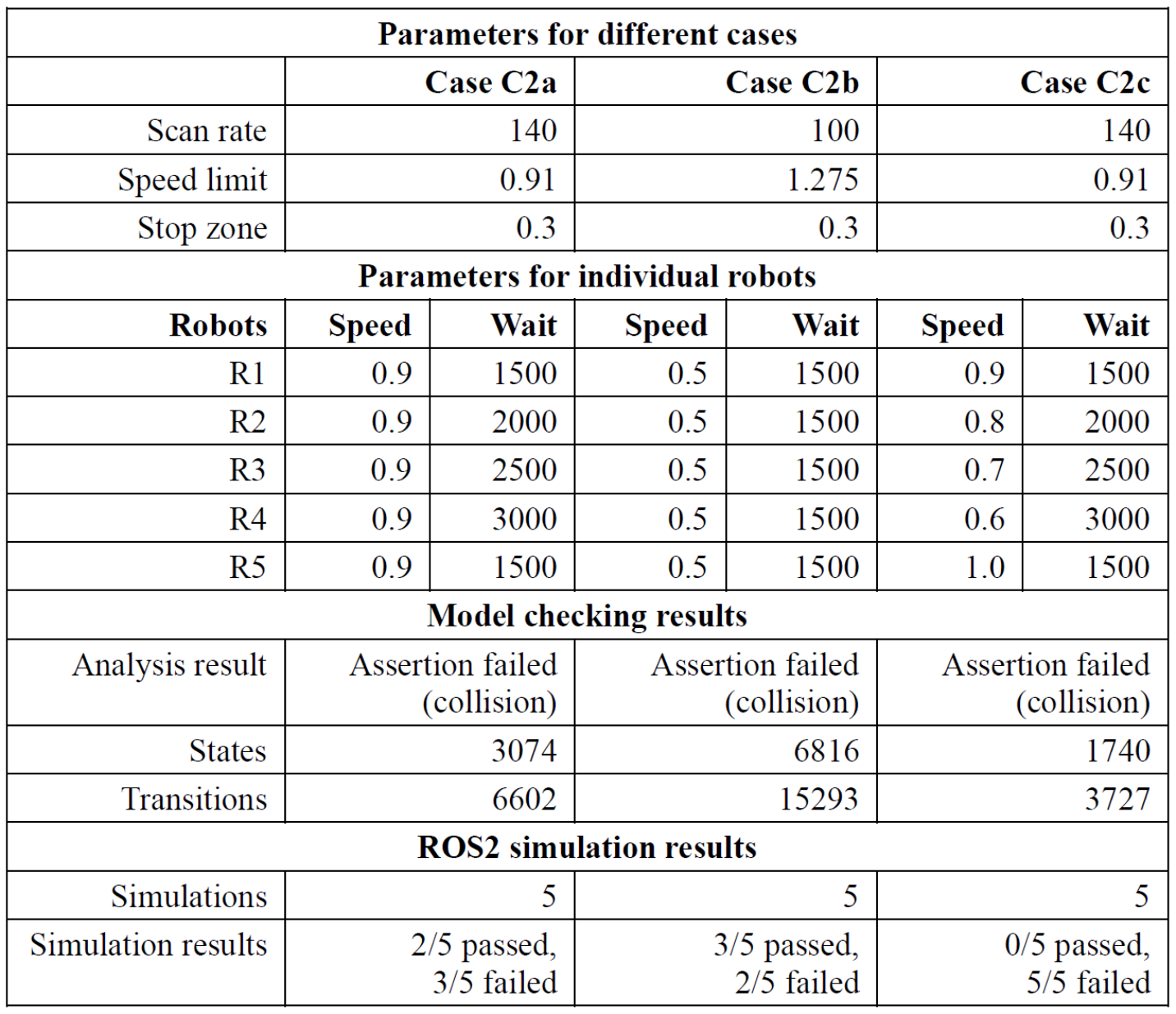}  
\end{tabular}%
}
\end{table}

\textbf{Non-working cases (Cases C2a, C2b, C2c).} 
Table \ref{table:4} provides the statistical and experimental parameters and results for cases C2a, C2b, C2c. Each of the three cases violates the safe operating constraints in a distinct way (speeds that violate the scan-rate constraint, short stop distance, and uniform speed parameters across all robots), which explains both why the model checker fails and why the ROS2 failure rates differ across cases.

C2a uses a scan rate of $140ms$ with all five robots running at the same speed of $0.9 m/s$. Applying the Nyquist-derived constraint: the maximum safe speed is $0.255 / (2 × 0.14) = 0.91 m/s$ — so $0.9 m/s$ is right at the boundary. However, the uniform speed across all robots means there is no natural staggering: robots tend to arrive at intersection points simultaneously, overwhelming the conflict resolution. The model checker finds a collision after only 3,074 states, and 3/5 ROS2 simulations fail.

C2b keeps the scan rate at $100ms$ and speed limit at $1.275 m/s$ — identical to the working cases globally — but assigns all robots the same speed ($0.5 m/s$) and crucially the same wait time ($1500ms$). This complete uniformity in both speed and wait time creates a worst-case scenario:
 robots move in lockstep, systematically converging on intersections together. The model checker needs slightly more exploration ($6,816$ states) before finding the collision, reflecting that the failure is less immediate than in C2a. $2/5$ simulations fail in ROS2.

C2c is the most aggressive case: scan rate $140ms$, speed limit $0.91$ m/s, with robot speeds ranging from $0.6$ to $1.0 m/s$. R5 at $1.0 m/s$ exceeds the Nyquist-derived safe limit of $0.91 m/s$, and even the lower-speed robots ($0.6\operatorname{-}0.9 m/s$) are operating at or near the boundary given the slower scan rate of $140ms$. The combination of a degraded scan rate and a wide speed spread means that faster robots outrun their sensing capability while slower robots still create congestion, producing conflicts across the entire robot team rather than just at one robot. This explains why C2c yields the fewest states ($1,740$) before a counterexample is found and the worst ROS2 outcome — $0/5$ simulations pass — as the failure conditions are pervasive across all robots rather than dependent on a specific unlucky timing.

The state counts in the non-working cases ($1,740\operatorname{-}6,816$) are dramatically lower than in the working cases ($15,747\operatorname{-}20,554$). This is because the model checker terminates as soon as it finds a counterexample. A bad configuration leads to a violation quickly, so only a small portion of the state space is explored before the search stops. In contrast, a safe configuration requires exhaustive exploration of the entire reachable state space before correctness can be declared. The low state counts in C2a–C2c are therefore a direct indicator of how early and how easily the system fails. The ROS2 simulations reproduce the same failures, confirming that the model's predictions are faithfully transferred to the real system.

Even in the non-working cases, some ROS2 simulations pass (2/5 in C2a, 3/5 in C2b). It reflects the non-deterministic nature of concurrent execution. The model checker proves that a collision is reachable, not that it is inevitable in every run. Whether a collision actually manifests depends on the precise timing of scheduler decisions at runtime. C2c's 0/5 pass rate shows that when the Nyquist analogous constraint (Equation~\ref{eq:2}) is directly violated, the failure becomes effectively unavoidable.

\begin{figure}[tbp]
    \centering
    \includegraphics[width=\columnwidth,keepaspectratio]{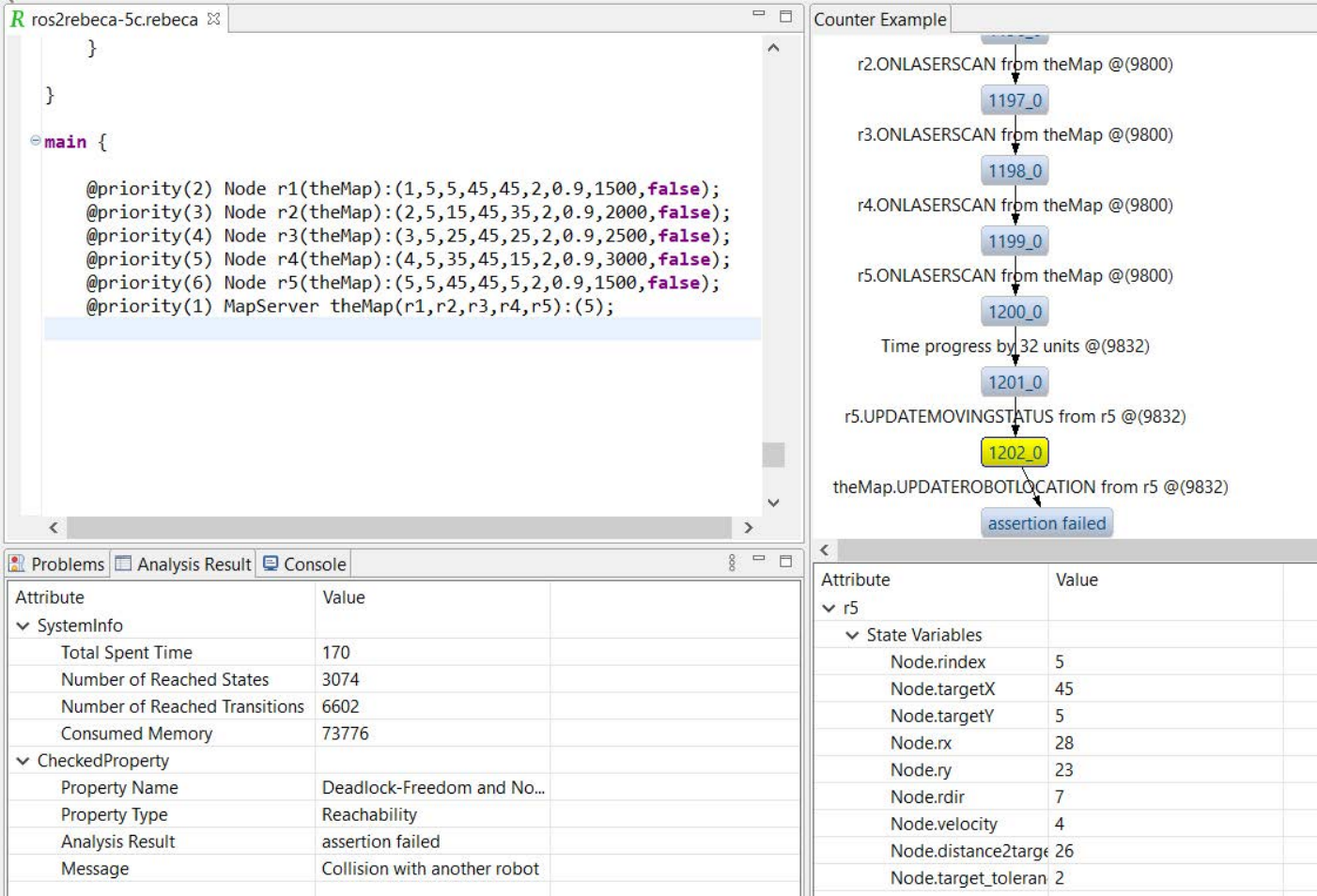}
    \includegraphics[width=\columnwidth,,keepaspectratio]{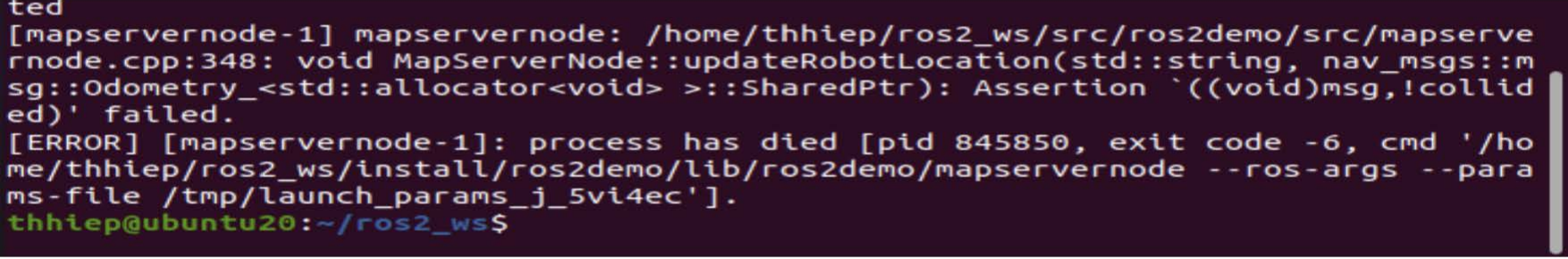}
    \includegraphics[width=\columnwidth,keepaspectratio]{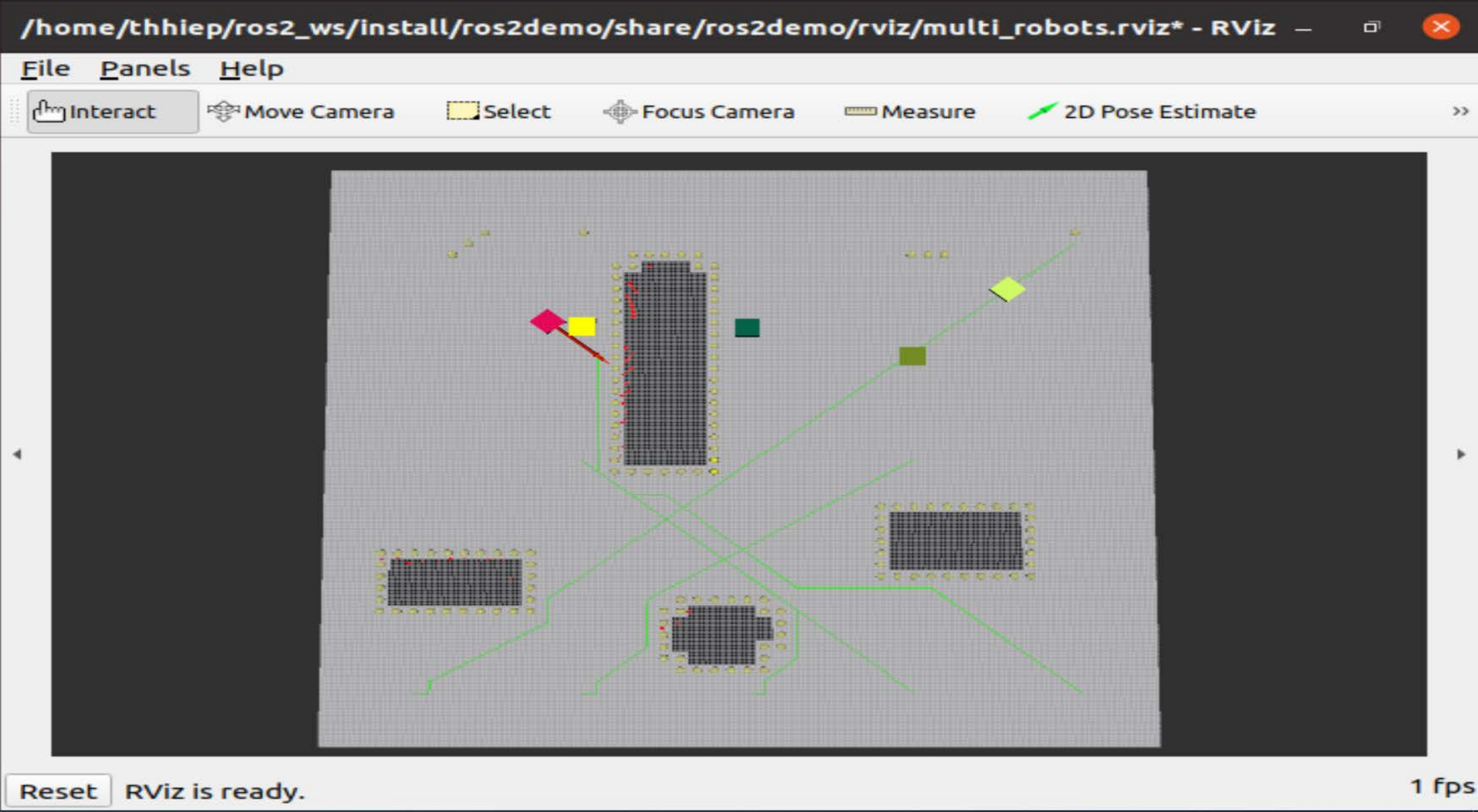}
    \caption{
    Screenshots of Afra results (up) and ROS2 simulation results (down) for
    a non-working case}
    \label{fig:nonworking}
\end{figure}
Figure~\ref{fig:nonworking} illustrates a representative non-working case. The counterexample trace (upper-right panel) shows the exact sequence of events leading to the collision: laser scan messages arriving at timestamps 9800–9832ms, followed by R5 updating its moving status and then reporting its location to the map server, at which point the assertion fails with the message "Collision with another robot". The state variables of R5 at the moment of failure are also shown — position ($rx=28$, $ry=23$), direction 7, velocity 4, and distance to target 26 — pinpointing precisely where and when the collision occurs in the model.
The terminal output (middle-bottom panel) confirms the same failure in ROS2: the map server node crashes with an assertion error — "Collision with another robot" — and the process exits with code -6, indicating an abnormal termination triggered by the same condition the model predicted.

The RViz2 visualization (bottom panel) shows the physical consequence: two robots (red and yellow) have converged on the same location near an obstacle, with their overlapping positions visible in the top-left area of the map. The frame rate has dropped to 1 fps, reflecting the system crash, compared to 5 fps in the working case.

\section{Discussion, Limitations, and Future Work}\label{sec:discuss}
Our experimental results confirm that model checking accurately predicts the actual behavior of the system. A multi-robot system implemented as a concurrent multi-agent system cannot be fully verified through traditional simulation or real deployment alone --- concurrency and nondeterminism mean that not all scenarios can be verified through case-by-case testing. Model-checking addresses this by systematically exploring the entire reachable state space. 

However, model checking introduces its own challenges: the system must be expressed as a discrete, abstracted model to remain tractable. For an autonomous multi-robot system with continuous physical behaviors, this raises two fundamental questions. First, how to model the system and the interactions discretely and compactly while still correctly capturing the relevant behaviors? Second, how to ensure that the model faithfully represents the real system  so that conclusions drawn on the model are also valid on the real system in practice? These are the questions that motivated our work. Some prior work did not adequately address this discrete-continuous gap, resorting to integer-only variables rather than properly handling real-valued quantities. 

Robotic systems are complex by nature, even basic multi-robot systems already exhibit significant system-level complexity. Oversimplified models risk losing essential domain information. In particular, many existing works neglect geometric and kinematic features such as physical dimensions, speed, and motion constraints. The central challenge is therefore to capture the necessary physical complexity while retaining the simplicity and tractability required for model checking.

\textbf{Question 1: How do you model a robotic system discretely and compactly while still correctly capturing the relevant behaviors?}

The key is to properly account for the physical constraints of the system rather than relying on naive or pattern-based randomization. Robotic systems are constrained by kinematic features, geometry, and physics laws — if these are not incorporated, the model generates infeasible scenarios that cause unbounded state space explosion. In this work, the discretization strategy explicitly handles the discrete-continuous gap through the Nyquist-derived speed constraints and grid-based abstractions, ensuring that the discrete model only explores physically realizable states. The model is intentionally kept as a baseline that captures movement, pathfinding, obstacle management, and robot interaction at a level of abstraction that generalizes beyond the specific robot platform.

\textbf{Question 2: How do you know the model faithfully represents the real system?}

The work addresses this through two complementary models serving distinct purposes. An initial prototype model is created according to the workflow diagram in Fig.~\ref{fig:workflow} prior to the code, to test architectural design and processing logic, discovering parameters that influence the behaviors of the system, anticipating potential issues to cope with, and experimenting behavioral algorithms. A separate verification model is then derived after the code is written, reflecting what has been implemented. The alignment between the verification model and the code is confirmed by the experimental results themselves: correctness properties verified in the verification model are consistently matched by the ROS2 simulation outcomes --- violations predicted by model checking reproduce as actual failures, and configurations verified as safe pass all simulation runs. While simulation produces fluctuating results due to concurrency and nondeterminism, model checking gives a consistent yes or no answer, making the correspondence between model and system unambiguous and repeatable.

Note that verifying distinct properties may require deriving separate verification models at varying levels of abstraction. A model A is an abstraction of model B if both represent the same system but A has less detail. A is a sound abstraction if every property of interest that holds in A also holds in B — equivalently, B is a refinement of A that adds detail without violating properties established by A. A is a complete abstraction if every property of interest that holds in B also holds in A. Both soundness and completeness are always defined with respect to the specific properties of interest, not the model as a whole.
In this work, we use Timed Rebeca for developing both the initial prototype model and the verification model for the purpose of verifying reachability, deadlock freedom, and collision freedom with the abstractions described in Section~\ref{sec:model} and ~\ref{sec:challenge}, reducing the discrete-continuous gap. This approach integrates prototyping and verification within a
unified framework.

The faithfulness of the Timed Rebeca model to the ROS2 implementation rests on three explicit geometric and temporal conditions. First, 
the robot's position is represented by its frame origin (at its center), which is mapped to the center of the nearest grid cell as the robot moves between cells. The robot's physical dimensions are accounted for by projecting its footprint into the continuous world when checking for collisions. Equation~\ref{eq:4} ensures that this footprint projection is fine-grained enough that any collision occurring during travel between two adjacent cells in the continuous world is guaranteed to be detected in the discrete model --- no collision can be missed by the cell-level abstraction. Second, the stop zone must exceed the edge length ($0.3m > 0.255m$), ensuring a robot is never simultaneously in two cells. Third, the speed must satisfy the Nyquist-derived constraint (Equation~\ref{eq:2}), ensuring the robot's position is observable at cell-level granularity at every scan instant. When these three conditions hold, every concrete transition in the ROS2 system has a corresponding transition in the discrete model, and the safety properties — collision freedom and deadlock freedom — are preserved across the abstraction. The experimental results in Section 8 provide empirical confirmation: all property violations found in the model reproduced as failures in ROS2, and all verified-safe configurations passed all simulation runs.

\paragraph{\textbf{Limitations}}
There are several limitations in this work:
\begin{itemize}
    \item The codes of Timed Rebeca model and ROS2 implementation are manually constructed and synchronized. It is not straightforward and error-prone to map from a discretely abstracted model to real-world implementation code with many details, and vice versa. They all need to be derived from a common higher-level model for semantic conformance and synchronization.
    
    \item Semantic preservation was verified manually by selective spot tests with representative cases in safe or unsafe areas of the parameter space and manual comparison of traces and outcomes.
    
    \item \textbf{\textit{Scalability}.} Our experiments can handle 5 robots in a 50x50 grid, which is already a larger system compared to others' work (e.g. by Gu et al. \cite{10.1145/3193992.3193999} which can only handle one robot with one simple moving obstacle). The 50x50 grid size was chosen to set a limit on the worst-case path length (which can be 50x50) and to remain manageable to Rebeca model checker.
    In order to check the scalability, we obtained verification results by running Afra on a grid of 100x100 and 10 robots keeping their path contained in a 50x50 grid. The experiment was conducted on an HP Pavilion+ laptop (Core i7, Windows 11) equipped with a 477 GB hard disk and 16 GB of RAM. Afra generated 434,615 states and 2,151,259 transitions.
    If we change the configuration by increasing the length of the path or the number of robots, the model-checking process continues until all available memory is exhausted. 
    
    The bottleneck for state explosion is the storage of the full generated path in the state space, whereas the path length on an $m\times m$ grid in worst case can be $m \times m$. The routing algorithm and data structures can be adjusted to tackle the state space explosion problem.     

    \item \textbf{\textit{Resolution vs. accuracy.}} The relation between the resolution and accuracy of the discrete model is captured by Equation \ref{eq:4} through analysis and experimentation, not fully empirically evaluated. Through experimentation, we detected that the model starts to behave correctly when Equation \ref{eq:4} is satisfied. If the resolution is reduced beyond that limit, that means the model is too coarse/sparse, it will result in uncaptured important moments, as stated. Conversely, if the  resolution is increased, meaning the step size is thinner than necessary, by analysis, Equation \ref{eq:4} is still respected, safety properties are still preserved as before, resulting in no gain in accuracy -- that's why we did not go further than that limit. Doubling resolution means halving the step size, thus doubling the grid dimensions, which is equal to the problem of scalability assessment.
\end{itemize}

\paragraph{\textbf{Future Work}}

Currently, both the Timed Rebeca model and the ROS2 implementation are manually constructed and semantically synchronized, with abstraction soundness verified through manual spot checks. A key future direction is to introduce a continuous model as a higher-level representation from which both the discrete model-checking model and the implementation code are generated automatically, with semantic preservation formally proved as part of the transformation process. This would simultaneously eliminate manual synchronization effort, shield developers from discretization complexity, and replace selective manual validation with systematic, provable soundness guarantees across the entire parameter space.

A complementary direction is the design of a unified language that encompasses both discrete and continuous aspects within a single framework, enabling automated generation of verification models and ROS2 code from a common specification. This requires formalizing the mapping rules between modelling and implementation constructs, enhancing the discretization strategies developed in this work, and establishing quantitative metrics for evaluating any remaining model–system gaps.

\section{Related Work}\label{sec:cmp}
This section reviews existing works on modelling and formal verification of robotic systems, with a particular focus on ROS-based and multi-robot systems, to contextualise and substantiate the contributions of this work.

Halder et al. \cite{10.5555/3101290.3101303} used Timed Automata and UPPAAL to model and detect communication issues in a ROS pub-sub channel, focusing on queue overflow and message interruptions from higher priority publishers. The analysis was demonstrated on the Kobuki robot, verifying only concurrency and timing issues in a communication thread, without addressing robotic behaviors. 

Gu et al. \cite{10.1145/3193992.3193999} used Timed Automata to model the control system of an autonomous wheel loader, including algorithms for path planning (A*) and collision avoidance (Dipole Flow Field), and used UPPAAL to encode and model-check against several requirements via TCTL logic. The verification properties include the following: compute collision-free paths given a dummy mobile obstacle (moving along a preset path), switch to stopping mode upon malfunction, and complete a cruise within a time limit. The solution works with a single vehicle, moving objects are treated as dimensionless dots, LIDAR data is skipped, and dynamic path planning has issues with arbitrarily moving obstacles. Difficulty in using integer variables in UPPAAL to model robotic behaviors was mentioned but not addressed, and no comparison with actual system behaviors was provided. 

Drona \cite{drona} is a framework for reliable planning of collision-free 3D flying paths for a swarm of drones with consideration of local clock deviations. Drona uses the P language for modelling, then map the code in P to a deployable ROS application in C language. The P code of Drona can be verified using the Zing model checker for P programs. The multi-robot motion planner in Drona can compute timed trajectories for all drones so that they fly coordinatedly in time without collision. Drona specifically works with connected drones, assuming each fits one grid cell. Our solution works for dimensional ground robots and arbitrary moving obstacles dynamically detected by a LIDAR sensor, and each robot behaves independently and adaptively up to its own situation.

In \cite{10.1007/s10270-018-00710-z} and \cite{10.1007/s10514-024-10163-7}, the authors present RoboChart, a domain-specific modelling language based on UML state machine diagram supported by the RoboTool environment built on the Eclipse Modelling Framework/Ecore to model robot control software. The toolchain generates both deployable code and mathematical models, enabling automated verification of correctness properties expressed in a natural language like assertion syntax. The formal foundation is discrete-timed CSP, with FDR as the accompanied model checker. The consistency between the program and the model is verified by recording the execution trace in the program and feed it back to the model checker to verify if the model passes it as well. The generated robot code is a custom C++ framework rather than ROS. Physical aspects concerning the robot and its interactions with
the environment are not modelled \cite{10.1007/s10514-024-10163-7}. The approach  focuses on software modelling only. It is not clear how to map ROS2 system building blocks like the map data, the nodes and their interactions, simulated sensory data, planning algorithms and movement/action trajectories to the RoboChart language and tooling. Modelling of continuous behaviours is left for a future work.

Mohammed et al. \cite{10.5772/7349} used hybrid automata and hybrid state machine to specify and model-check a rescue scenario in a multi-robot system called RoboCup. However, as experienced with Hybrid Rebeca and hybrid automata~\cite{hybridrebeca}, it is not straightforward to express system architecture using hybrid automata. The analysis also suffers from usability and scalability.

In \cite{10.1007/978-3-031-35257-7_1}, Nigam and Talcott present automated methods to prove safety properties for cyber physical systems, which involves steps of reading sensors and  sampling time between observations. The adequacy of sampling time step is addressed so that nothing important is missed between steps. It is a fixed rate sampling approach which is known to suffer from rounding-off and performance problems (the rate at some point cannot get smaller because pseudo-real number representation in computers has limited precision, and the number of iterations is huge). The way we model laser scans is analogous to their approach (scanning step is fixed to $2\textsuperscript{o}$), while the way we model the movements can be considered as discrete-event simulation, which captures representative points and models the time advancement to the next point. In any case, the step length in terms of time or distance has to be small enough to avoid missing adversarial events that may happen between two points.

Balabonski et al. \cite{DBLP:conf/netys/BalabonskiCPRTU19} study formal models of mobile robot movement under different assumptions about time and synchronization:
(i) robots operating in a continuous 2D Euclidean space, moving in continuous time with asynchronous progress (continuous models), and (ii) robots abstracted to move on discrete structures (e.g., graphs) with discrete asynchronous steps (discrete models).
The continuous model captures physical reality (real positions and motions) while the discrete model captures the digital nature of many algorithmic robot protocols --- where moves are abstracted as atomic asynchronous steps. The work uses the Pactole framework and Coq proof assistant to specify the two theoretical models and prove that any execution in one model can be simulated in the other, thus certifying that a discrete reasoning framework is safely applicable to continuous real robots. Their work uses abstract mathematical models of asynchronous robot movements and provides formal reasoning about equivalence between continuous and discrete models. 
Zilio et al.~\cite{DALZILIO2023104301} proposed a framework that generates executable models from formal specifications, enabling runtime monitoring and verification while preserving semantic consistency between specification and execution. Their framework allows properties such as schedulability, timing constraints, and mutual exclusion to be verified offline using model-checking techniques while also supporting runtime safety mechanisms. 
Praveen et al.~\cite{Praveen9734238} proposed a design and formal verification framework to model multi-agent protocols, focusing on a distributed drone scenario. Their approach represents the system as networks of timed automata in UPPAAL and translates the verified models into ROS-based simulations to enable early validation before deployment.
Model-driven approaches have also been explored for analysing human–robot interaction scenarios. In particular, prior work~\cite{Lestingi22}  extends a model-driven development framework with statistical model checking to estimate mission success probabilities in multi-robot service scenarios, incorporating richer models of human behaviour and task handover among robots in complex healthcare environments.

In contrast, our work is a more engineering-oriented, pragmatic approach, addressing the bridging between a realistic ROS2-based multi-robot system and a discrete system model which is based on the actor model for asynchronous interactions and uses timing constructs in Timed Rebeca to approximate continuous physical behavior. 

\section{Conclusion}\label{sec:conclude}
Robotic systems are inherently complex and span multiple disciplines. As these systems grow in size and sophistication, modelling them prior to implementation becomes both intuitive and highly beneficial. Model-based development and verification enable early-stage prototyping and more efficient problem resolution. In this work, we have demonstrated a concrete methodology for co-designing and verifying multi-robot systems using Timed Rebeca, whose asynchronous, message-based execution model and explicit time semantics make it naturally suited for representing ROS2 node topologies and performing model-checking to validate essential system properties.

Realizing this advantage entails systematically handling the intricate aspects of robotics: from bridging the gap between the continuous nature of physical-world dynamics and the discrete abstractions in formal models, to resolving discrepancies between real-world programming environments and modelling languages. This work has successfully tackled these challenges by constructing a precise, efficient model of a multi-robot system and using that model to guide ROS2 software development. Three critical system properties were verified --- target reachability, collision freedom, and deadlock freedom --- with experimental results confirming that property violations detected in the model consistently reproduce as failures in the ROS2 implementation, and that configurations verified as safe pass all simulation runs. 

Thus, the central technical contribution is a principled strategy for bridging the discrete-continuous gap --- translating continuous robot dynamics into discrete models that are compact enough for model checking yet faithful enough that verified properties carry over to the real implementation.

\section*{Acknowledgement}
This work was partially funded by the Swedish Research Council through the ORPHEUS project (rn. 2022-03408).
\nopagebreak
\FloatBarrier
\bibliographystyle{elsarticle-num}
\bibliography{ros2rebeca-arxiv-fixed}

@article{DBLP:journals/fuin/SirjaniMSB04,
  author       = {Marjan Sirjani and
                  Ali Movaghar and
                  Amin Shali and
                  Frank S. de Boer},
  title        = {Modeling and Verification of Reactive Systems using Rebeca},
  journal      = {Fundam. Informaticae},
  volume       = {63},
  number       = {4},
  pages        = {385--410},
  year         = {2004},
  url          = {http://content.iospress.com/articles/fundamenta-informaticae/fi63-4-05},
  bibsource    = {dblp computer science bibliography, https://dblp.org}
}

@inproceedings{DBLP:conf/fmco/Sirjani06,
  author       = {Marjan Sirjani},
  editor       = {Frank S. de Boer and
                  Marcello M. Bonsangue and
                  Susanne Graf and
                  Willem P. de Roever},
  title        = {Rebeca: Theory, Applications, and Tools},
  booktitle    = {Formal Methods for Components and Objects, 5th International Symposium,
                  {FMCO} 2006, Amsterdam, The Netherlands, November 7-10, 2006, Revised
                  Lectures},
  series       = {Lecture Notes in Computer Science},
  volume       = {4709},
  pages        = {102--126},
  publisher    = {Springer},
  year         = {2006},
  url          = {https://doi.org/10.1007/978-3-540-74792-5\_5},
  doi          = {10.1007/978-3-540-74792-5\_5},
  bibsource    = {dblp computer science bibliography, https://dblp.org}
}

@inproceedings{DBLP:conf/fsen/KhamespanahSK23,
  author       = {Ehsan Khamespanah and
                  Marjan Sirjani and
                  Ramtin Khosravi},
  editor       = {Hossein Hojjat and
                  Erika {\'{A}}brah{\'{a}}m},
  title        = {Afra: An Eclipse-Based Tool with Extensible Architecture for Modeling
                  and Model Checking of Rebeca Family Models},
  booktitle    = {Fundamentals of Software Engineering - 10th International Conference,
                  {FSEN} 2023, Tehran, Iran, May 4-5, 2023, Revised Selected Papers},
  series       = {Lecture Notes in Computer Science},
  volume       = {14155},
  pages        = {72--87},
  publisher    = {Springer},
  year         = {2023},
  url          = {https://doi.org/10.1007/978-3-031-42441-0\_6},
  doi          = {10.1007/978-3-031-42441-0\_6},
  bibsource    = {dblp computer science bibliography, https://dblp.org}
}

@article{ref1,
  author       = {Hong Hiep Trinh},
  title        = {Model based development and verification of {ROS2} robotic applications using {Timed Rebeca}, {Master Thesis}, {Malardalen University}},
  year         = {2023},
}

@book{ref8,
  author    = {Edmund M. Clarke and
               Orna Grumberg and
               Daniel Kroening and
               Doron A. Peled and
               Helmut Veith},
  title     = {Model Checking},
  edition   = {2},
  publisher = {MIT Press},
  address   = {Cambridge, MA, USA},
  year      = {2018},
  isbn      = {978-0-262-03883-6},
  url       = {https://mitpress.mit.edu/books/model-checking-second-edition}
}

@article{ref11,
    author = {{Open Robotics}},
    title = {{ROS2 online documentation}},
    url = "https://docs.ros.org/en/foxy/index.html",
    urldate = "2023-12-29" ,
    year = "2023",
}

@article{ref12,
    author = {{Open Robotics}},
    title = {{Understanding ROS2 nodes}},
    url = "https://docs.ros.org/en/foxy/Tutorials/Beginner-CLI-Tools/Understanding-ROS2-Nodes/Understanding-ROS2-Nodes.html",
    urldate = "2023-12-29",
    year = "2023",
}

@article{ref13,
    author = {{Open Robotics}},
    title = {{ROS2 Concepts}},
    url = "https://docs.ros.org/en/foxy/Concepts.html",
    urldate = "2023-12-29",
    year = "2023",
}

@article{ref14,
    author = {{OSRF}},
    title = {{ROS Concepts and Design Patterns}},
    url = "https://osrf.github.io/ros2multirobotbook/ros2_design_patterns.html",
    urldate = "2023-12-29",
    year = "2023",
}

@article{ref15,
title = {Formal Specification and Verification of Autonomous Robotic Systems: A Survey},
author = {Luckcuck, Matt and Farrell, Marie and Dennis, Louise A. and Dixon, Clare and Fisher, Michael},
year = {2019},
issue_date = {September 2020},
publisher = {Association for Computing Machinery},
address = {New York, NY, USA},
volume = {52},
number = {5},
issn = {0360-0300},
doi = {10.1145/3342355},
journal = {ACM Comput. Surv.},
month = {sep},
articleno = {100},
numpages = {41},
}

@masterthesis{ref17,
    author = "Filip Berlin and Sebastian Granath",
    title = "Obstacle Detection and Avoidance for an Automated Guided Vehicle",
    school = "Linköping University" ,
    year = 2021,
    url = "https://urn.kb.se/resolve?urn=urn:nbn:se:liu:diva-177709"
}

@ARTICLE{ref18,
  author={Shannon, C.E.},
  journal={Proceedings of the IRE}, 
  title={Communication in the Presence of Noise}, 
  year={1949},
  volume={37},
  number={1},
  pages={10-21},
  doi={10.1109/JRPROC.1949.232969}}

@article{ref19,
    author = "Wikipedia",
    title = "A* search algorithm",
    url = "https://en.wikipedia.org/wiki/A*_search_algorithm",
    urldate = "2023-12-29",
    year = "2023",
}

@ARTICLE{ref19b,
  author={Hart, Peter E. and Nilsson, Nils J. and Raphael, Bertram},
  journal={IEEE Transactions on Systems Science and Cybernetics}, 
  title={A Formal Basis for the Heuristic Determination of Minimum Cost Paths}, 
  year={1968},
  volume={4},
  number={2},
  pages={100-107},
  doi={10.1109/TSSC.1968.300136}}

@article{ref21,
title = {Approaches of Road Boundary and Obstacle Detection Using LIDAR},
author = {O. Yalcin and A. Sayar and O.F. Arar and S. Akpinar and S. Kosunalp},
year = {2013},
note = {IFAC Workshop on Advances in Control and Automation Theory for Transportation Applications},
}

@inproceedings{10.1145/3193992.3193999,
author = {Gu, Rong and Marinescu, Raluca and Seceleanu, Cristina and Lundqvist, Kristina},
title = {Formal Verification of an Autonomous Wheel Loader by Model Checking},
year = {2018},
isbn = {9781450357180},
publisher = {Association for Computing Machinery},
address = {New York, NY, USA},
url = {https://doi.org/10.1145/3193992.3193999},
doi = {10.1145/3193992.3193999},
pages = {74–83},
numpages = {10},
location = {Gothenburg, Sweden},
series = {FormaliSE '18}
}

@article{10.1145/3542945,
author = {Araujo, Hugo and Mousavi, Mohammad Reza and Varshosaz, Mahsa},
title = {Testing, Validation, and Verification of Robotic and Autonomous Systems: A Systematic Review},
year = {2023},
issue_date = {March 2023},
publisher = {Association for Computing Machinery},
address = {New York, NY, USA},
volume = {32},
number = {2},
issn = {1049-331X},
url = {https://doi.org/10.1145/3542945},
doi = {10.1145/3542945},
journal = {ACM Trans. Softw. Eng. Methodol.},
month = {mar},
articleno = {51},
numpages = {61},
}

@inproceedings{10.5555/3101290.3101303,
author = {Halder, Raju and Proença, José and Macedo, Nuno and Santos, André},
title = {Formal Verification of ROS-Based Robotic Applications Using Timed-Automata},
year = {2017},
isbn = {9781538604229},
publisher = {IEEE Press},
booktitle = {Proceedings of the 5th International FME Workshop on Formal Methods in Software Engineering},
pages = {44–50},
numpages = {7},
location = {Buenos Aires, Argentina},
series = {FormaliSE '17}
}

@inproceedings{DBLP:conf/birthday/Sirjani18,
  author       = {Marjan Sirjani},
  title        = {Power is Overrated, Go for Friendliness! Expressiveness, Faithfulness,
                  and Usability in Modeling: The Actor Experience},
  booktitle    = {Principles of Modeling - Essays Dedicated to Edward A. Lee on the
                  Occasion of His 60th Birthday},
  series       = {Lecture Notes in Computer Science},
  volume       = {10760},
  pages        = {423--448},
  publisher    = {Springer},
  year         = {2018},
  url          = {https://doi.org/10.1007/978-3-319-95246-8\_25},
}

@article{DBLP:journals/scp/ReynissonSACJIS14,
  author       = {Arni H. Reynisson and
                  Marjan Sirjani and
                  Luca Aceto and
                  Matteo Cimini and
                  Ali Jafari and
                  Anna Ing{\'{o}}lfsd{\'{o}}ttir and
                  Steinar H. Sigurdarson},
  title        = {Modelling and simulation of asynchronous real-time systems using {Timed
                  Rebeca}},
  journal      = {Sci. Comput. Program.},
  volume       = {89},
  pages        = {41--68},
  year         = {2014},
url          = {https://doi.org/10.1016/j.scico.2014.01.008},
}

@inproceedings{DBLP:conf/compsac/SirjaniLK20,
  author       = {Marjan Sirjani and
                  Edward A. Lee and
                  Ehsan Khamespanah},
  title        = {Model Checking Software in Cyberphysical Systems},
  booktitle    = {44th {IEEE} Annual Computers, Software, and Applications Conference,
                  {COMPSAC} 2020, Madrid, Spain, July 13-17, 2020},
  pages        = {1017--1026},
  publisher    = {{IEEE}},
  year         = {2020},
url          = {https://doi.org/10.1109/COMPSAC48688.2020.0-138},
}

@article{brown2004,
  author  = {Brown, Alan},
  title   = {Model Driven Architecture: Principles and Practice},
  journal = {Software and Systems Modeling},
  volume  = {3},
  number  = {4},
  pages   = {314--327},
  year    = {2004},
  doi     = {10.1007/s10270-004-0061-2}
}

@book{fishwick2007,
  editor       = {Paul A. Fishwick},
  title        = {Handbook of Dynamic System Modeling - Chapter 11},
  publisher    = {Chapman and Hall/CRC},
  year         = {2007},
  url          = {https://doi.org/10.1201/9781420010855},
  doi          = {10.1201/9781420010855},
  isbn         = {978-1-58488-565-8},
  bibsource    = {dblp computer science bibliography, https://dblp.org}
}

@book{Agha1986,
  author    = {Gul Agha},
  title     = {Actors: A Model of Concurrent Computation in Distributed Systems},
  publisher = {MIT Press},
  address   = {Cambridge, MA, USA},
  year      = {1986},
  isbn      = {0262010925}
}

@INPROCEEDINGS{drona,
  author={Desai, Ankush and Saha, Indranil and Yang, Jianqiao and Qadeer, Shaz and Seshia, Sanjit A.},
  booktitle={2017 ACM/IEEE 8th International Conference on Cyber-Physical Systems (ICCPS)}, 
  title={DRONA: A Framework for Safe Distributed Mobile Robotics}, 
  year={2017},
  volume={},
  number={},
  pages={239-248},
  doi={}
}

@inproceedings{DBLP:journals/corr/abs-2309-07302,
  author       = {Marjan Sirjani and
                  Ehsan Khamespanah},
  editor       = {Claudio Antares Mezzina and
                  Georgiana Caltais},
  title        = {Timed Actors and Their Formal Verification},
  booktitle    = {Procs of {EXPRESS/SOS} 2023},
  series       = {{EPTCS}},
  volume       = {387},
  pages        = {1--7},
  year         = {2023},
  url          = {https://doi.org/10.4204/EPTCS.387.1},
  doi          = {10.4204/EPTCS.387.1},
  bibsource    = {dblp computer science bibliography, https://dblp.org}
}

@phdthesis{ehsanphd,
  title        = {Modeling, verification, and analysis of timed actor-based models},
  author       = {Khamespanah, Ehsan},
  year         = 2018,
  month        = {June},
  address      = {Reykjavik, Iceland},
  note         = {https://hdl.handle.net/20.500.11815/1185},
  school       = {Reykjavik University},
  type         = {PhD thesis}
}

@article{street2023multirobot,
author = {Street, Charlie and Mansouri, Masoumeh and Lacerda, Bruno},
year = {2023},
month = {08},
pages = {1-10},
title = {Formal Modelling for Multi-Robot Systems Under Uncertainty},
volume = {4},
journal = {Current Robotics Reports},
doi = {10.1007/s43154-023-00104-0}
}

@inproceedings{10.1145/2414639.2414645,
author = {Khamespanah, Ehsan and Sabahi Kaviani, Zeynab and Khosravi, Ramtin and Sirjani, Marjan and Izadi, Mohammad-Javad},
title = {Timed-rebeca schedulability and deadlock-freedom analysis using floating-time transition system},
year = {2012},
isbn = {9781450316309},
publisher = {Association for Computing Machinery},
address = {New York, NY, USA},
doi = {10.1145/2414639.2414645},
booktitle = {Proceedings of the 2nd Edition on Programming Systems, Languages and Applications Based on Actors, Agents, and Decentralized Control Abstractions},
pages = {23–34},
numpages = {12},
location = {Tucson, Arizona, USA},
series = {AGERE! 2012}
}

@inbook{10.5772/7349,
author = {Mohammed, Ammar and Furbach, Ulrich and Stolzenburg, Frieder},
year = {2010},
month = {01},
pages = {},
title = {Multi-Robot Systems: Modeling, Specification, and Model Checking},
isbn = {978-953-307-036-0},
doi = {10.5772/7349}
}

@inbook{hybridrebeca,
author = {Jahandideh, Iman and Ghassemi, Fatemeh and Sirjani, Marjan},
year = {2019},
month = {07},
pages = {3-27},
title = {Hybrid Rebeca: Modeling and Analyzing of Cyber-Physical Systems},
isbn = {978-3-030-23702-8},
doi = {10.1007/978-3-030-23703-5_1}
}

@InProceedings{10.1007/978-3-031-35257-7_1,
author="Nigam, Vivek
and Talcott, Carolyn",
editor="David, Cristina
and Sun, Meng",
title="Automating Recoverability Proofs for Cyber-Physical Systems with Runtime Assurance Architectures",
booktitle="Theoretical Aspects of Software Engineering",
year="2023",
publisher="Springer Nature Switzerland",
address="Cham",
pages="1--19",
isbn="978-3-031-35257-7"
}

@article{10.1007/s10270-018-00710-z,
author = {Miyazawa, Alvaro and Ribeiro, Pedro and Li, Wei and Cavalcanti, Ana and Timmis, Jon and Woodcock, Jim},
title = {{RoboChart: modelling and verification of the functional behaviour of robotic applications}},
year = {2019},
issue_date = {October   2019},
publisher = {Springer-Verlag},
address = {Berlin, Heidelberg},
volume = {18},
number = {5},
issn = {1619-1366},
url = {https://doi.org/10.1007/s10270-018-00710-z},
doi = {10.1007/s10270-018-00710-z},
journal = {Softw. Syst. Model.},
month = oct,
pages = {3097–3149},
numpages = {53}
}

@article{10.1007/s10514-024-10163-7,
author = {Li, Wei and Ribeiro, Pedro and Miyazawa, Alvaro and Redpath, Richard and Cavalcanti, Ana and Alden, Kieran and Woodcock, Jim and Timmis, Jon},
title = {{Formal design, verification and implementation of robotic controller software via RoboChart and RoboTool}},
year = {2024},
issue_date = {Aug 2024},
publisher = {Kluwer Academic Publishers},
address = {USA},
volume = {48},
number = {6},
issn = {0929-5593},
url = {https://doi.org/10.1007/s10514-024-10163-7},
doi = {10.1007/s10514-024-10163-7},
journal = {Auton. Robots},
month = jul,
numpages = {22}
}

@article{DBLP:journals/firai/FoughaliZ22,
  author       = {Mohammed Foughali and
                  Alexander Zuepke},
  title        = {Formal Verification of Real-Time Autonomous Robots: An Interdisciplinary
                  Approach},
  journal      = {Frontiers Robotics {AI}},
  volume       = {9},
  pages        = {791757},
  year         = {2022},
  url          = {https://doi.org/10.3389/frobt.2022.791757},
  doi          = {10.3389/FROBT.2022.791757},
  bibsource    = {dblp computer science bibliography, https://dblp.org}
}

@inproceedings{DBLP:conf/netys/BalabonskiCPRTU19,
  author       = {Thibaut Balabonski and
                  Pierre Courtieu and
                  Robin Pelle and
                  Lionel Rieg and
                  S{\'{e}}bastien Tixeuil and
                  Xavier Urbain},
  editor       = {Mohamed Faouzi Atig and
                  Alexander A. Schwarzmann},
  title        = {Continuous vs. Discrete Asynchronous Moves: {A} Certified Approach
                  for Mobile Robots},
  booktitle    = {Networked Systems - 7th International Conference, {NETYS} 2019, Marrakech,
                  Morocco, June 19-21, 2019, Revised Selected Papers},
  series       = {Lecture Notes in Computer Science},
  volume       = {11704},
  pages        = {93--109},
  publisher    = {Springer},
  year         = {2019},
  url          = {https://doi.org/10.1007/978-3-030-31277-0\_7},
  doi          = {10.1007/978-3-030-31277-0\_7},
  bibsource    = {dblp computer science bibliography, https://dblp.org}
}

@article{DALZILIO2023104301,
title = {A formal toolchain for offline and run-time verification of robotic systems},
journal = {Robotics and Autonomous Systems},
volume = {159},
pages = {104301},
year = {2023},
issn = {0921-8890},
doi = {https://doi.org/10.1016/j.robot.2022.104301},
url = {https://www.sciencedirect.com/science/article/pii/S0921889022001907},
author = {Silvano {Dal Zilio} and Pierre-Emmanuel Hladik and Félix Ingrand and Anthony Mallet}
}

@inproceedings{Lestingi22,
author = {Lestingi, Livia and Sbrolli, Cristian and Scarmozzino, Pasquale and Romeo, Giorgio and Bersani, Marcello M. and Rossi, Matteo},
title = {Formal modeling and verification of multi-robot interactive scenarios in service settings},
year = {2022},
isbn = {9781450392877},
publisher = {Association for Computing Machinery},
address = {New York, NY, USA},
url = {https://doi.org/10.1145/3524482.3527653},
doi = {10.1145/3524482.3527653},
booktitle = {Proceedings of the IEEE/ACM 10th International Conference on Formal Methods in Software Engineering},
pages = {80–90},
numpages = {11},
location = {Pittsburgh, Pennsylvania},
series = {FormaliSE '22}
}

@ARTICLE{Praveen9734238,
  author={Praveen, Adithya T. and Gupta, Anubhav and Bhattacharyya, Siddhartha and Muthalagu, Raja},
  journal={IEEE Systems Journal}, 
  title={Assuring Behavior of Multirobot Autonomous Systems With Translation From Formal Verification to ROS Simulation}, 
  year={2022},
  volume={16},
  number={3},
  pages={5092-5100},
  doi={10.1109/JSYST.2022.3149677}}
\end{document}